\documentclass[lettersize,journal]{IEEEtran}

\usepackage{color}
\usepackage{graphicx}

\usepackage{amsmath,amsfonts, amssymb}
\usepackage{array}
\usepackage{textcomp}
\usepackage{stfloats}
\usepackage{url}
\usepackage{verbatim}
\usepackage{caption}
\usepackage{subcaption}
\usepackage{cite}
\usepackage{cleveref}
\usepackage{multirow}
\usepackage{tabularx}
\usepackage{pifont}
\usepackage{makecell}
\crefname{figure}{Fig.}{Figs.}
\Crefname{figure}{Fig.}{Figs.}
\crefname{table}{Table}{Tables}
\Crefname{table}{Table}{Tables}

\crefname{equation}{Eq.}{Eqs.}
\Crefname{equation}{Eq.}{Eqs.}

\usepackage{booktabs}
\usepackage{enumitem}
\usepackage{bm}
\usepackage{siunitx}

\usepackage{amsmath,amsfonts}
\usepackage{algorithmic}
\usepackage{array}
\usepackage[caption=false,font=normalsize,labelfont=sf,textfont=sf]{subfig}
\usepackage{textcomp}
\usepackage{stfloats}
\usepackage{url}
\usepackage{verbatim}
\usepackage{graphicx}
\def\BibTeX{{\rm B\kern-.05em{\sc i\kern-.025em b}\kern-.08em
    T\kern-.1667em\lower.7ex\hbox{E}\kern-.125emX}}
\usepackage{balance}

\begin{document}
\title{Phase-Gated Constrained Reinforcement Learning for End-to-End Capture and Detumbling of Non-Cooperative Targets}
\author{Álvaro Belmonte-Baeza$^{*1}$, Celia Redondo-Verdú$^{2}$, José L. Ramón$^{2}$, Miguel Cazorla$^{1}$, \\Leonard Felicetti$^{3}$, and Jorge Pomares$^{2}$
\thanks{This research received funding from the European Union's European Innovation Council (EIC) Pathfinder programme under Grant Agreement No. 101223360 for the DEXTER project. Álvaro Belmonte-Baeza was supported by the Spanish Ministry of Universities under grant FPU21/02586.}
\thanks{$^{1}$Álvaro Belmonte-Baeza and Miguel Cazorla are with the Department of Computer Science and Artificial Intelligence, University of Alicante, Spain. *Corresponding email:
        {\tt\small alvaro.belmonte@ua.es}}%
\thanks{$^{2}$Celia Redondo-Verdú, José L. Ramón, and Jorge Pomares are with the Department of Physics, Systems Engineering and Signal Theory, University of Alicante, Spain.
       }%
\thanks{$^{3}$ Leonard Felicetti is with the Centre for Space Systems, Cranfield University, Cranfield, UK.}
\thanks{Code link: Project code will be added here upon acceptance to respect the double-blind review process. }
\thanks{This paper has supplementary downloadable material provided by the
authors, consisting of a video that illustrates the approach, capture and
detumbling behaviour.}
}

\markboth{}%
{Belmonte-Baeza \MakeLowercase{\textit{et al.}}: Phase-Gated Constrained Reinforcement Learning for End-to-End Capture and Detumbling}

\maketitle

\begin{abstract}
Active debris removal requires a servicer spacecraft to approach a non-cooperative tumbling object, capture it, and arrest its rotation. This sequence is conventionally divided among phase-specific controllers, each depending on knowledge of the target's inertia, a dedicated grappling fixture, or a hand-designed switch at the contact-rich seam where such pipelines are most fragile. We show that a single constrained reinforcement learning policy can execute the entire mission, driving a dual-arm free-flying servicer from free-space closing through friction-based fixture-free capture to full detumbling of a target whose mass and inertia are unknown and whose state is observed only through noisy estimates. The enabling mechanism is a smooth proximity gate applied at three loci of a constrained Markov decision process. In the reward it induces the phase structure rather than imposing it; in the action space it forecloses a degenerate grab-and-pull exploit as a hard guarantee; and in the constraint set it resolves a normalization failure of constraint-based learning, in which a constraint spanning phases that load it unequally is rendered nearly inert in the phase where precision matters most. Both the failure mode and the remedy apply to any multi-phase problem of this kind. Over $1000$ randomized trials the policy succeeds in $99.5\%$ while holding its contact-force, actuator, base-reaction and collision limits throughout, transfers zero-shot to unseen geometries and centre-of-mass offsets, and extrapolates across a tenfold range of target mass.

\end{abstract}



\section{Introduction}\label{sec:intro}

The sustained growth of orbital activity has left near-Earth space increasingly congested with non-functional satellites, spent rocket bodies, and fragmentation debris. Beyond the immediate collision risk to operational assets, this population is self-amplifying: each hypervelocity collision generates further fragments, driving the cascading instability first predicted by Kessler and Cour-Palais that threatens the long-term usability of key orbits \cite{kessler1978collision}. Active Debris Removal (ADR) has consequently become a mission priority \cite{shan2016review, ledkov2022review}, with in-orbit demonstrations such as \textit{RemoveDebris} beginning to validate candidate capture technologies \cite{aglietti2020removedebris}. The highest-value targets (large, massive, rapidly tumbling objects) are precisely those that are non-cooperative, lacking communication, attitude control, and the standardized grappling interfaces on which cooperative servicing relies \cite{fallahiarezoodar2025review, floresabad2014review}.
 
For this class of target, dual-arm space manipulator systems mounted on a free-flying servicer are widely regarded as the most capable solution, since they permit deliberate, distributed force application throughout contact \cite{papadopoulos2021survey, floresabad2014review}. The removal operation is conventionally decomposed into three phases: a pre-capture \emph{approach} that brings the end-effectors into position relative to the tumbling target, contact establishment and \emph{capture}, and post-capture \emph{stabilization} that dissipates the coupled system's residual momentum \cite{floresabad2014review}. Each phase is individually demanding. The approach requires tracking a tumbling target with a free-flying base whose attitude is itself perturbed by every manipulator motion \cite{papadopoulos1993singularities}; contact injects severe, unmodeled impact transients as the dynamic model changes discontinuously at grasp \cite{wu2018contact, dong2020combined}; and stabilization must arrest a large, uncertain inertia without saturating actuators or destabilizing the servicer's own attitude.
 
Model-based control has addressed these phases with considerable success when they are treated \emph{individually}: detumbling through optimal, adaptive, and fuel-efficient formulations \cite{li2024timeoptimal, zhou2025fuelsaving}, and capture through robust and impedance-based schemes for non-cooperative spinning targets \cite{wang2023robust, han2022capture}. These methods, however, share structural dependencies that limit deployment against real debris. They require accurate identification of the target's mass and inertia, a process that converges only after a dedicated estimation transient and that model-based offline trajectory optimization can ill afford \cite{dong2026detumbling}; many assume a dedicated grappling fixture on the target \cite{wang2023robust}; and they rely on hand-crafted logic to switch between phase-specific controllers. At the transition between phases, where the assumed dynamics change discontinuously at the instant of impact, no mechanism guarantees that safety limits on contact force, actuator effort, and base reaction are maintained uniformly. The transition is left as an open seam.
 
Reinforcement learning (RL) \cite{sutton2018reinforcement} is an attractive alternative for the regime that undermines model-based control: the severe base–manipulator dynamic coupling and inertial uncertainty of free-floating manipulation. Yet learning-based results in space robotics have so far addressed the mission one phase at a time: pre-capture guidance and
motion planning \cite{hovell2021deep,cao2023rlprior,wang2022learning}, the physical
capture or soft-capture grasp \cite{beigomi2024softcapture}, or post-capture
detumbling from an already-secured grasp \cite{dong2026detumbling}. Each
typically assumes a favorable initial configuration in which the hardest parts
of the phase transitions are bypassed. Moreover, standard RL offers no mechanism to strictly enforce safety: reward shaping, however carefully tuned, cannot prevent a policy from selecting constraint-violating actions at deployment \cite{achiam2017cpo}, where a single excessive contact force or a lost grasp can end the mission.
 
Constrained RL (CRL) addresses this last point by casting the problem as a Constrained Markov Decision Process (CMDP), expressing safety as bounds on cumulative cost rather than as soft penalties \cite{altman1999constrained}. Among CRL solvers, the Constraints-as-Terminations (CaT) formulation reframes each constraint violation as a stochastic episode termination, embedding safety enforcement directly into the discounted policy-gradient objective with negligible overhead \cite{chanesane2024cat}. Constraint-based learning of this kind has recently produced high-performing, safe controllers for legged locomotion \cite{kim2024constraints, chanesane2024cat} and, in a robot-manipulation setting, has been given a rigorous safe-action-space treatment \cite{liu2024constraintmanifold}. Its application to orbital manipulation, where contact-force, actuator-saturation, base-reaction, and collision limits must hold \emph{simultaneously and continuously} through contact-rich maneuvers and across phase boundaries, remains unexplored.
 
This leaves a clear gap: no single learned controller executes the complete ADR mission as one policy, spanning approach, fixture-free capture and full detumbling with safety enforced
continuously across the phase transitions. The difficulty is not merely additive. The terminal approach and the capture--stabilization phase pursue fundamentally conflicting objectives: the former demands collision-free closing and precise pose alignment with a tumbling target, holding the manipulators quiescent against a moving base, while the latter demands contact-rich, force-regulated energy dissipation with those same manipulators. Naively summing phase-specific rewards yields a policy that learns neither cleanly. The central technical challenge is therefore a reward and integration design under which a \emph{single} policy discovers all three behavioral modes and sequences them safely, with no hand-designed switching.

We address this challenge with a CRL framework for end-to-end ADR on a dual-arm servicer equipped with disk-shaped end-effectors and a 6-DoF propulsion system. A single policy drives the servicer from a contact-free initial configuration to complete stabilization of the coupled system through friction-based pressure capture of a featureless tumbling target of unknown mass and inertia. Because the target's inertial parameters are unobserved and its state is available only through noisy estimates \cite{sharma2020pose, napolano2025cnn}, the task is naturally a partially observable decision problem \cite{lauri2023pomdp} in which the policy must adapt online rather than rely on identified parameters. Safety is enforced throughout via a newly proposed multi-stage extension of the CaT formulation \cite{chanesane2024cat}. Training and evaluation are conducted in a high-fidelity, GPU-parallel rigid-body physics environment \cite{makoviychuk2021isaacgym, mittal2025isaaclab} that resolves the contact dynamics, actuator saturation, and continuously shifting composite center of mass that idealized numerical models omit.
 
To the best of our knowledge, this is the first learning-based controller to execute the complete approach--capture--stabilization sequence for a non-cooperative tumbling target as a single policy, without fixtures on the
target, prior identification of its inertial parameters, or hand-designed switching between phase-specific controllers. The mechanism that makes this possible is a single, smoothly varying phase signal applied consistently at three distinct loci of the CMDP: the reward, the action space, and the constraint set. The main contributions of this paper are:
 
\begin{itemize}
    \item \textbf{End-to-end learned ADR as a single policy.} We formulate the full approach--capture--stabilization mission against a tumbling target of unknown inertia as one CRL problem and solve it with a single policy, removing the fixtures, inertial-parameter identification, and phase-switching logic on which prior pipelines depend.

    \item \textbf{Gated reward composition for conflicting phase objectives.} Approach and capture demand opposite behaviors of the same manipulators. A gate and its complement form a partition of unity interpolating between holding the arms quiescent during closing and tracking a co-rotating target face during capture, while a second, contact-derived gate cascades onto the stabilization terms so that detumbling reward is reachable only through a well-formed grasp. Naive summation fails to learn either behavior cleanly; the gated policy self-organizes into approach, capture, and detumbling modes with smooth transitions across the contact boundary.

    \item \textbf{Action-space gating as a hard guarantee.} The same gate multiplies the manipulator command, suppressing arm motion at range. This \emph{forecloses} the degenerate solution in which the arms seize the target and draw it toward the base instead of the base translating toward the target, a behavior reward shaping can penalize but not rule out. Ablation confirms the grasp becomes unreliable only when this guarantee and the reward gating are both removed.

    \item \textbf{Constraint decomposition for multi-phase constrained RL.} We identify a failure mode of CaT~\cite{chanesane2024cat}: because each constraint is normalized by a single scale tracking its largest violation anywhere in the task, a constraint spanning several phases is rendered nearly inert in the phase where precision matters most. We resolve it by splitting each such constraint into an always-active loose term and a contact-gated tight term, each with its own scale. Ablation shows the natural single-constraint alternative fails as predicted, and the remedy applies to any multi-phase CRL problem.

    \item \textbf{Comprehensive evaluation in high-fidelity simulation.} We report whole-task and per-phase performance over $1000$ trials; zero-shot transfer to unmodeled geometries and center-of-mass offsets; mass and tumbling-rate extrapolation that localizes the binding difficulty of fixture-free capture; ablations isolating each gating locus and the decomposition; and a comparison against representative model-based controllers.
\end{itemize}

The remainder of this paper is organized as follows. \Cref{sec:related_work} reviews model-based and learning-based approaches to ADR and situates the contribution. \Cref{sec:problem} formalizes the servicer, the target, the contact model, and the assumptions under which the controller operates. \Cref{sec:preliminaries} recalls the constrained RL formulation and the Constraints-as-Terminations algorithm. \Cref{sec:method} presents the gating
mechanism and the constraint design that constitute the core of this work. \Cref{sec:results} describes the experimental setup and reports the results, and \cref{sec:conclusions} discusses limitations and directions for future work. A video summarizing the method and results accompanies this paper.
\section{Related Work}\label{sec:related_work}

\subsection{Model-Based Capture and Detumbling of Non-Cooperative Targets}

Robotic removal of non-cooperative targets has been studied extensively under a model-based lens, and comprehensive surveys trace the maturation of capture mechanisms and control strategies \cite{shan2016review, floresabad2014review, papadopoulos2021survey}. The post-capture stabilization problem, in which a chaser must dump the residual momentum of a tumbling target while preserving its own attitude, has been addressed through time-optimal and coordinated trajectory optimization \cite{li2024timeoptimal}, fuel-efficient detumbling for dual-arm closed-chain systems \cite{zhou2025fuelsaving}, and momentum-redistribution schemes that channel the combined angular momentum into reaction wheels \cite{luo2020detumbling}. The capture problem has been treated with robust sliding-mode and impedance controllers designed to reject the model uncertainty of a spinning target \cite{wang2023robust, han2022capture}, and friction-based detumbling that regulates the tangential contact force to spin down the target while holding the base attitude has been demonstrated for dual-arm servicers \cite{chen2020friction}. A recurring feature of these works is a dependence on the target's inertial parameters. Because these are unknown for real debris, they must be identified online or compensated adaptively: adaptive and task-compatibility controllers estimate the inertia during detumbling \cite{zhou2023adaptive}, and optimal concurrent formulations fold actuator and target-motion bounds into the control law while adapting to the same  uncertainty \cite{xi2024adaptive}. In every case the compensation converges only after a dedicated estimation transient, which renders offline model-based trajectory optimization operationally impractical for time-critical maneuvers \cite{dong2026detumbling, wang2024positionforce}. These are precisely the crutches that the present work removes.



\subsection{Learning-Based Control for Space Manipulation}

Reinforcement learning has been applied to free-floating and free-flying space robots to exploit its capacity for handling nonlinear, strongly coupled dynamics without an explicit model \cite{BELMONTE2026}. A first body of work targets the approach phase: RL has been used for spacecraft proximity-operations guidance \cite{hovell2021deep}, and for motion planning of dual-arm free-floating manipulators toward non-cooperative objects, including tracking targets that spin at unknown rates \cite{cao2023rlprior, wang2022learning}. Recent work has combined RL with Lyapunov-based control for proximity operations \cite{holt2025rl}. A second body of work targets the capture phase, where RL with tactile feedback has been shown to improve soft-capture success against free-floating debris \cite{beigomi2024softcapture}. Further research targets post-capture detumbling, where RL controllers stabilize a captured target with uncertain inertia and adapt across initial conditions without retraining \cite{dong2026detumbling}. Experimental studies on air-bearing testbeds have validated learned guidance for the pre-grasp motion, but they still hand off to a separately optimized law for the manipulator deployment \cite{hovell2022laboratory}. In every case, the learned policy addresses a single phase and assumes that the preceding phases have delivered a favorable initial configuration. The present work departs from this pattern, learning the entire
approach--capture--stabilization sequence within one policy and under a constrained formulation that enforces safety throughout. Broader surveys of RL for spacecraft control corroborate both the promise of the paradigm and the open challenges of high-fidelity training and controller verification that any such method must confront \cite{tipaldi2022rl}.

\subsection{End-to-End Versus Modular Policies for Long-Horizon Contact-Rich Control}

Outside the space domain, the trade-off between end-to-end and modular learned control has been examined directly for long-horizon, contact-rich tasks. Modular architectures that decompose a task into semantic sub-skills connected by a learned or scripted selector perform well within each sub-skill but tend to accumulate errors at the boundaries between them, where the local assumptions of adjacent controllers meet \cite{sun2025arch}. Hierarchical hybrid schemes mitigate this by pairing a library of specialized primitives with a high-level selector \cite{sun2025arch}; such designs, however, reintroduce a discrete switching mechanism whose transitions must themselves be managed. A single policy that internalizes the phase structure avoids an explicit handoff and preserves global task context across the transition, at the cost of a substantially harder learning problem when the constituent phases pursue conflicting objectives. Our staged reward and integration design is aimed squarely at making that harder problem tractable, and Section VI quantifies, through ablation, the failure of naive reward composition that motivates it.The staged-reward formulation is not specific to ADR; it applies to any multi-phase task of this kind.

\subsection{Safe and Constrained Reinforcement Learning}

Enforcing hard limits on a learned controller is the subject of the CRL literature. The Constrained MDP provides the formal setting \cite{altman1999constrained}, and Constrained Policy Optimization (CPO) introduced a trust-region method with near-constraint-satisfaction guarantees at each update \cite{achiam2017cpo}. Subsequent methods trade off strictness, tunability, and computational cost: Lagrangian relaxations of policy-gradient methods \cite{ray2019benchmarking}, interior-point formulations that impose logarithmic barriers on cumulative constraints \cite{liu2020ipo}, exact-penalty variants that avoid second-order computation \cite{zhang2022p3o}, and recovery-based schemes that separate task and safety policies for contact-rich manipulation \cite{thananjeyan2021recovery}. Recent work has placed safe RL on a constraint manifold with formal safe-action-space guarantees and demonstrated it on physical robots \cite{liu2024constraintmanifold}. Within this landscape, CaT is attractive for orbital manipulation because it enforces constraints in a state-wise, per-step fashion by construction \cite{chanesane2024cat}: whereas episodic, cumulative-constraint methods bound expected cost over a trajectory, they cannot guarantee satisfaction at the single instant of impact, which is exactly where an ADR controller is most vulnerable. Enforcing the same contact-force, actuator, base-reaction, and collision constraints continuously across all three phases is a central requirement of our formulation. The standard CaT formulation, however, does not transfer cleanly to multi-phase tasks whose phases load a constraint at different magnitudes. \Cref{sec:method:cat_gating} introduces phase-gated constraints that apply only where they are needed, and the remedy carries over to any constrained multi-phase setting.
%
%

\section{Problem Formulation}
\label{sec:problem}

This section defines the servicer, the target, and the contact model that
together specify the active debris removal (ADR) task, states the sensing
assumptions under which the controller must operate, and formalises the
resulting decision problem.

\subsection{Servicer Dynamics}
\label{sec:problem:servicer}

The servicer is a free-flying spacecraft equipped with two $7$-degree-of-freedom
(DoF) manipulators, each terminated by a flat, disk-shaped end effector. Let
$\mathcal{I}$ denote an inertial frame, $\mathcal{B}$ the servicer base frame,
and $\mathcal{T}$ the target body frame. The base pose is
$(\bm{p}_b,\bm{R}_b)\in\mathbb{R}^3\times SO(3)$ with twist
$\bm{v}_b=[\dot{\bm{p}}_b^{\top},\,\bm{\omega}_b^{\top}]^{\top}\in\mathbb{R}^6$,
and the joint configuration is $\bm{q}\in\mathbb{R}^{14}$.

The equations of motion of the combined system are
\begin{equation}
\label{eq:eom}
\begin{bmatrix}
\bm{H}_{b} & \bm{H}_{bm}\\[2pt]
\bm{H}_{bm}^{\top} & \bm{H}_{m}
\end{bmatrix}
\begin{bmatrix}
\dot{\bm{v}}_{b}\\[2pt]
\ddot{\bm{q}}
\end{bmatrix}
+
\begin{bmatrix}
\bm{c}_{b}\\[2pt]
\bm{c}_{m}
\end{bmatrix}
=
\begin{bmatrix}
\bm{f}_{b}\\[2pt]
\bm{\tau}
\end{bmatrix}
+
\sum_{i\in\{\mathrm{L},\mathrm{R}\}}
\begin{bmatrix}
\bm{J}_{b,i}^{\top}\\[2pt]
\bm{J}_{m,i}^{\top}
\end{bmatrix}
\bm{f}_{c,i},
\end{equation}
where $\bm{H}_{b}\in\mathbb{R}^{6\times 6}$ and
$\bm{H}_{m}\in\mathbb{R}^{14\times 14}$ are the base and manipulator inertia
matrices, $\bm{c}_{b}$ and $\bm{c}_{m}$ collect Coriolis and centrifugal terms,
$\bm{\tau}\in\mathbb{R}^{14}$ are the joint torques, and $\bm{f}_{c,i}$ is the
contact wrench at end effector $i$, mapped onto the base and the joints by the
Jacobians $\bm{J}_{b,i}$ and $\bm{J}_{m,i}$.

The off-diagonal block $\bm{H}_{bm}$ is the dynamic coupling between base and
manipulators, and the origin of the principal difficulty addressed here: it is
configuration dependent, changes in effect once contact is established, and gives
rise to dynamic singularities whose location cannot be predicted from the
manipulator kinematics alone~\cite{papadopoulos1993singularities}. 

The base wrench $\bm{f}_{b}=[\bm{f}_{\mathrm{thr}}^{\top},\,
\bm{\tau}_{\mathrm{rw}}^{\top}]^{\top}$ is produced by a $6$-DoF propulsion
system and follows the commanded wrench through first-order dynamics
\begin{equation}
\label{eq:actuator_lag}
\tau_{a}\,\dot{\bm{f}}_{b} + \bm{f}_{b} = \bm{f}_{b}^{\mathrm{cmd}},
\qquad \tau_{a} = 0.05~\mathrm{s},
\end{equation}
subject to per-axis authority limits on thrust and reaction torque. The
manipulator joints are torque limited to $39$ and $13~\mathrm{N}\,\mathrm{m}$ at
the proximal and distal joints, with velocity limits of $1.39$ and
$1.22~\mathrm{rad}\,\mathrm{s}^{-1}$ respectively.

\subsection{Target Model}
\label{sec:problem:target}

The target is a non-cooperative, uncontrolled rigid body: it carries no
grappling fixture, no docking interface, and no fiducial markers, and it does
not communicate. It is modelled as a cube of side $0.25~\mathrm{m}$ with mass $m_{t}$ and inertia
tensor $\bm{I}_{t}$, neither known to the controller. In the absence of contact
its dynamics are those of a freely tumbling body,
\begin{align}
\label{eq:target_dyn}
m_{t}\dot{\bm{v}}_{t} &= \textstyle\sum_{i} \bm{f}_{c,i}, \\
\bm{I}_{t}\dot{\bm{\omega}}_{t} + \bm{\omega}_{t}\times\bm{I}_{t}\bm{\omega}_{t}
&= \textstyle\sum_{i}\left(\bm{r}_{c,i}\times\bm{f}_{c,i} + \bm{\tau}_{c,i}\right),
\end{align}
with $\bm{f}_{c,i}$, $\bm{\tau}_{c,i}$ the force and moment applied by end effector
$i$ at contact point $\bm{r}_{c,i}$ about the target centre of mass. Before
contact both sums vanish and the gyroscopic term
$\bm{\omega}_{t}\times\bm{I}_{t}\bm{\omega}_{t}$ drives a nutating motion whose
axis is not fixed in the body frame. Each component of
$\bm{\omega}_{t}(0)$ is drawn uniformly up to $0.25~\mathrm{rad}\,\mathrm{s}^{-1}$
and $m_t$ over $[15,45]~\mathrm{kg}$, so the servicer contends with a target
whose inertia varies threefold across episodes without being told which instance
it faces.

\subsection{Fixture-Free Capture}
\label{sec:problem:contact}

Capture is achieved by pressing the two disk-shaped end effectors against
opposing faces of the target: the grasp is sustained by friction alone, with no
form closure and no mechanical interlock. Decomposing the contact force at end
effector $i$ into normal and tangential components with respect to the contacted
face,
$\bm{f}_{c,i} = f_{n,i}\bm{n}_{i} + \bm{f}_{t,i}$, the contact sticks only while
\begin{equation}
\label{eq:friction_cone}
\lVert \bm{f}_{t,i}\rVert \;\le\; \mu\, f_{n,i},
\qquad f_{n,i} \ge 0,
\end{equation}
where $\mu$ is the friction coefficient of the contact pair. Condition
\eqref{eq:friction_cone} distinguishes fixture-free capture from a fixture-based
grasp and drives the design that follows. The grasp exists only while actively maintained, since releasing the normal force releases the target. The normal force is bounded below by the tangential load needed to detumble the unknown inertia, and above by the structural limits of what is defunct hardware rather than a purpose-built interface. Capture therefore lives inside a force corridor
that must be found without knowledge of the inertial parameters and respected at
every instant, not on average. Because the arms press as an opposed pair their
normal forces largely cancel at the base; any imbalance injects a net wrench into
\eqref{eq:eom} through $\bm{J}_{b,i}^{\top}\bm{f}_{c,i}$ and disturbs the base
attitude, which the servicer must simultaneously regulate.

\subsection{Mission Definition and Success Criteria}
\label{sec:problem:mission}

The mission begins with the servicer holding station at a stand-off distance
from a freely tumbling target, and is conventionally decomposed into three
phases:

\begin{enumerate}
\item \textbf{Terminal approach.} The servicer closes the separation
      $d = \lVert \bm{p}_{t} - \bm{p}_{b}\rVert$ under thruster control with the
      manipulators held clear, arresting its motion at a stand-off from which the
      target faces lie within the workspace.
\item \textbf{Capture.} The end effectors press on two opposing faces until a
      friction grasp is established, without imparting an impulse that ejects the
      target or destabilises the base.
\item \textbf{Detumbling.} The angular momentum of the coupled system is
      dissipated to rest while the grasp and the base attitude are held.
\end{enumerate}

\begin{figure*}[t]\centering
  \includegraphics[width=\linewidth]{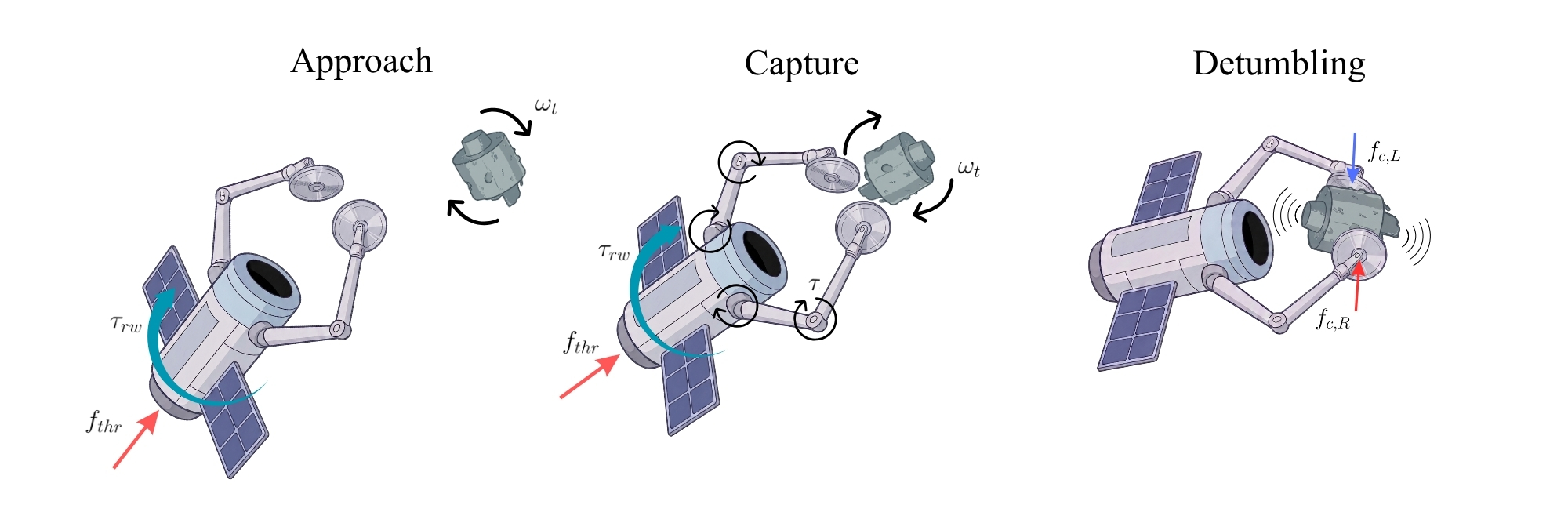}
  \caption{\textbf{The three phases of the ADR mission}. The servicer closes the
  separation under thruster force $\bm{f}_{\mathrm{thr}}$ and reaction torque
  $\bm{\tau}_{\mathrm{rw}}$ with the arms stowed; the arms are then actuated by
  joint torques $\bm{\tau}$ to press two disk-shaped end effectors on opposing
  faces of the tumbling target, and the resulting contact wrenches
  $\bm{f}_{c,\mathrm{L}}$, $\bm{f}_{c,\mathrm{R}}$ dissipate its angular momentum
  $\bm{\omega}_{t}$. The phases are named for exposition only: no phase index is
  supplied to the controller and no switching condition is evaluated at run time.}
  \label{fig:problem_overview}
\end{figure*}

These names serve exposition and the per-phase analysis of
Section~\ref{sec:results} only; they are not exposed to the controller. No phase
index enters the observation, no phase-dependent controller is selected, and no
switching condition is evaluated at run time: the policy must discover the phase
structure, and the transitions between phases, from the task objective alone. An illustration of the described task is shown in \cref{fig:problem_overview}

An episode is deemed successful if, within the mission horizon $T$, the servicer
brings the target to rest,
\begin{equation}
\label{eq:success}
\lVert\bm{v}_{t}\rVert \le \varepsilon_{v},
\qquad
\lVert\bm{\omega}_{t}\rVert \le \varepsilon_{\omega},
\end{equation}
with $\varepsilon_{v}=0.01~\mathrm{m}\,\mathrm{s}^{-1}$ and
$\varepsilon_{\omega}=0.03~\mathrm{rad}\,\mathrm{s}^{-1}$, while maintaining
contact with both end effectors and violating no critical constraint
(\cref{sec:method:constraints}). The angular tolerance is set deliberately above
the constraint limit imposed during learning
($0.02~\mathrm{rad}\,\mathrm{s}^{-1}$): the policy is driven toward a residual
stricter than the rate against which it is scored, so that reported success
reflects a margin rather than the boundary of the optimised objective.

\subsection{Sensing and Observation Model}
\label{sec:problem:sensing}

The controller does not observe the true target state. It is assumed that an
upstream relative-navigation front end supplies estimates of the target's
relative pose and twist, corrupted by zero-mean Gaussian noise:
\begin{align}
\label{eq:noise}
\tilde{\bm{p}}_{t}^{b} &= \bm{p}_{t}^{b} + \bm{\eta}_{p},
  & \bm{\eta}_{p} &\sim \mathcal{N}(\bm{0},\sigma_{p}^{2}\bm{I}_{3}), \nonumber\\
\tilde{\bm{R}}_{t}^{b} &= \bm{R}_{t}^{b}\,\mathrm{Exp}(\bm{\eta}_{\theta}),
  & \bm{\eta}_{\theta} &\sim \mathcal{N}(\bm{0},\sigma_{\theta}^{2}\bm{I}_{3}), \nonumber\\
\tilde{\bm{v}}_{t}^{b} &= \bm{v}_{t}^{b} + \bm{\eta}_{v},
  & \bm{\eta}_{v} &\sim \mathcal{N}(\bm{0},\sigma_{v}^{2}\bm{I}_{3}), \nonumber\\
\tilde{\bm{\omega}}_{t}^{b} &= \bm{\omega}_{t}^{b} + \bm{\eta}_{\omega},
  & \bm{\eta}_{\omega} &\sim \mathcal{N}(\bm{0},\sigma_{\omega}^{2}\bm{I}_{3}).
\end{align}
Attitude noise is applied as a right-multiplied rotation through the exponential
map, so that the perturbed estimate remains on $SO(3)$; the noisy rotation is
only subsequently encoded for the policy (Section~\ref{sec:method:spaces}).

The noise magnitudes are not free parameters but are set consistently with
reported relative-navigation accuracy on non-cooperative tumbling targets:
learned pose networks achieve centimetre-level position and degree-level attitude
on non-cooperative spacecraft~\cite{sharma2020pose}, filter-based trackers sustain
comparable accuracy on tumbling targets~\cite{park2023adaptive}, and ADR-specific
pipelines report angular rates accurate to a tenth of a degree per
second~\cite{napolano2025cnn}. The velocity figures in that work correspond to
acquisition at tens of metres; at the sub-metre separations considered here a
converged filter is substantially more accurate, and
$\sigma_{v}=5~\mathrm{mm}\,\mathrm{s}^{-1}$ is adopted
(Table~\ref{tab:noise}). This leaves $\sigma_{v}$ at half the success tolerance
$\varepsilon_{v}$ of \eqref{eq:success}: the policy must regulate the target to a
residual velocity of the same order as the uncertainty with which it measures
that velocity, and must therefore rely on the contact configuration and its own
proprioception, not the relative-state estimate alone, to certify that the target
has come to rest.

Proprioceptive quantities (joint positions and velocities, base twist, and
end-effector contact forces) are assumed measured onboard with negligible error
relative to the estimation uncertainty in \eqref{eq:noise}, as is standard for
encoder, inertial, and force/torque instrumentation.

\begin{table}[t]
\centering
\caption{\textbf{Relative-state estimation noise} (standard deviations).}
\label{tab:noise}
\begin{tabular}{llll}
\toprule
Quantity & Symbol & Value & Reference \\
\midrule
Relative position    & $\sigma_{p}$      & $1\times 10^{-2}~\mathrm{m}$                        & \cite{sharma2020pose} \\
Relative attitude    & $\sigma_{\theta}$ & $1.75\times 10^{-2}~\mathrm{rad}$ & \cite{sharma2020pose,park2023adaptive} \\
Relative linear vel. & $\sigma_{v}$      & $5\times 10^{-3}~\mathrm{m}\,\mathrm{s}^{-1}$       & \cite{napolano2025cnn} \\
Target angular vel.  & $\sigma_{\omega}$ & $1.7\times 10^{-3}~\mathrm{rad}\,\mathrm{s}^{-1}$ & \cite{napolano2025cnn} \\
\bottomrule
\end{tabular}
\end{table}

\subsection{Assumptions}
\label{sec:problem:assumptions}

The scope of the results that follow is delimited by the following assumptions.

\begin{enumerate}[label=\textbf{A\arabic*},leftmargin=*,itemsep=2pt]
\item \label{as:sensing}
      \textbf{Noisy relative state, no ground truth.} The relative pose and
      twist of the target are available only through the noisy estimates
      \eqref{eq:noise}. The task is therefore a partially observable decision
      problem~\cite{lauri2023pomdp}.

\item \label{as:inertia}
      \textbf{Unknown inertial parameters.} The mass and inertia tensor of the
      target are unknown and are never identified, online or offline. The
      controller must adapt to the instance it faces rather than adapting a
      model of it.

\item \label{as:scope}
      \textbf{Terminal-approach scope.} The mission considered here begins at a
      stand-off of approximately $1~\mathrm{m}$; the approach phase spans
      roughly $0.5~\mathrm{m}$ of closing motion. Long-range rendezvous and
      far-field relative orbit control are assumed to have been completed by an
      upstream guidance system and are outside the scope of this work. The
      target tumbles but does not translate relative to the servicer at the
      start of the episode.

\item \label{as:contact}
      \textbf{Rigid-body contact.} Both bodies are rigid. Contact is modelled
      with Coulomb friction and compliant normal response. Flexible appendages,
      fuel slosh, and structural deformation of the target are not modelled.

\item \label{as:orbit}
      \textbf{Microgravity, orbit-free.} Gravity is neglected. Over the
      $T=30~\mathrm{s}$ mission horizon, orbital effects (gravity-gradient torque,
      $J_2$, differential drag) are small compared with the contact and
      actuation wrenches and are omitted.
\end{enumerate}

Assumption~\ref{as:scope} most narrowly delimits the claims of this paper. The
contribution concerns the \emph{integration} of the terminal approach with
contact-rich capture and detumbling within a single policy; long-range
rendezvous guidance, for which mature solutions already exist, is outside it.

\subsection{Problem Statement}
\label{sec:problem:statement}

The control problem is now stated. Let $\bm{o}_{k}$ denote the observation
available at control step $k$, constructed from the noisy target estimates
\eqref{eq:noise} and the servicer's proprioception, and let $\bm{a}_{k}$ denote
the command applied to the manipulators and the propulsion system. Given a set
of safety constraints $\{c_{j}\}_{j=1}^{J}$, the objective is to find a single
policy $\pi:\; \bm{o}_{k} \;\longmapsto\; \bm{a}_{k}$ that drives an arbitrary instance of the task, from the contact-free initial configuration of Section~\ref{sec:problem:mission} to the success condition \eqref{eq:success}, satisfying every constraint $c_{j}$ at every step of every phase and across the transitions between them, without any externally supplied phase signal, without a grappling fixture, and without knowledge of the target's inertial parameters.

Two properties distinguish this problem from the sub-problems solved in prior work. First, the phases pursue objectives that are not merely different but \emph{opposed}: the approach requires the base to translate while the arms remain quiescent, whereas capture requires the base to hold station while the arms track and press a rotating face, so a reward that sums the two rewards a policy for doing each badly. Second, the constraint set is not phase invariant: the base velocity that constitutes safe, efficient closing during approach is a gross
violation of the reactionless-base requirement during pressing, so a single fixed limit is either too loose to protect the capture or too tight to permit the approach.

Section~\ref{sec:method} shows that both properties are addressed by the same
mechanism.
\section{Preliminaries}
\label{sec:preliminaries}

This section recalls the CRL formulation and the CaT algorithm~\cite{chanesane2024cat} on which the method of \cref{sec:method} is built. The material is standard, with the exception of the constraint-normalization property introduced in \cref{sec:prelim:cat}, on which the constraint design of \cref{sec:method:cat_gating} depends.

\subsection{Constrained Markov Decision Processes}
\label{sec:prelim:cmdp}

The task is modelled as a discounted MDP $(\mathcal{S},\mathcal{A},r,\gamma,\mathcal{T})$, in which RL seeks a policy $\pi$ maximizing the expected discounted return
\begin{equation}
\label{eq:rl_objective}
\max_{\pi}\;
\mathbb{E}_{\tau\sim\pi,\mathcal{T}}
\left[\sum_{t=0}^{\infty}\gamma^{t}\, r(\bm{s}_{t},\bm{a}_{t})\right].
\end{equation}
Because the servicer observes the target only through the noisy estimates of \eqref{eq:noise}, the true state is unavailable and the problem is formally a partially observable MDP~\cite{lauri2023pomdp}; as is standard in robot learning, it is addressed by providing the policy with a history-free observation augmented with the previous action.

Constrained RL augments the MDP with constraint functions $\{c_{i}:\mathcal{S}\times\mathcal{A}\rightarrow\mathbb{R},\; i\in I\}$, adopting the convention that constraint $i$ is \emph{violated} when $c_{i}(\bm{s},\bm{a})>0$, so that each $c_{i}$ is a signed violation margin: the amount by which a physical quantity exceeds its limit. The classical formulation bounds the discounted sum of constraint costs by a budget~\cite{altman1999constrained,achiam2017cpo}, but safety-critical tasks are better served by bounding the probability of violation at each step,
\begin{equation}
\label{eq:chance_constraint}
\mathbb{P}_{(\bm{s},\bm{a})\sim\rho_{\gamma}^{\pi,\mathcal{T}}}
\left[\,c_{i}(\bm{s},\bm{a})>0\,\right] \;\le\; \tilde{\epsilon}_{i},
\qquad \forall i\in I,
\end{equation}
where $\rho_{\gamma}^{\pi,\mathcal{T}}$ is the discounted occupancy induced by $\pi$. The distinction matters here: a budget permits the policy to spend its allowance where most profitable (e.g. exceeding the contact-force limit at the instant of impact and compensating thereafter) which is unacceptable when a single excessive force can fracture the target or eject it from the workspace.

\subsection{Constraints as Terminations}
\label{sec:prelim:cat}

CaT~\cite{chanesane2024cat} enforces \eqref{eq:chance_constraint} by converting violations into \emph{stochastic terminations of future reward}. Introducing $\delta_{t}\in[0,1]$, the objective becomes
\begin{equation}
\label{eq:cat_objective}
\max_{\pi}\;
\mathbb{E}_{\tau\sim\pi}
\left[\sum_{t=0}^{\infty}
\left(\prod_{t'=0}^{t}\gamma\bigl(1-\delta(\bm{s}_{t'},\bm{a}_{t'})\bigr)\right)
r(\bm{s}_{t},\bm{a}_{t})\right],
\end{equation}
so that violating a constraint attaches a probability of forfeiting \emph{all} subsequent reward. 

The termination probability is set proportional to the severity of the violation. With $c_{i}^{+}=\max(0,\,c_{i}(\bm{s},\bm{a}))$, the positive part of the violation margin,
\begin{equation}
\label{eq:cat_delta}
\delta \;=\; \max_{i\in I}\;
p_{i}^{\max}\,
\mathrm{clip}\!\left(\frac{c_{i}^{+}}{c_{i}^{\max}},\,0,\,1\right),
\end{equation}
where $p_{i}^{\max}\in(0,1]$ bounds the termination probability constraint $i$ can induce. Constraints with $p_{i}^{\max}=1$ are \emph{hard}; those with $p_{i}^{\max}<1$ are \emph{soft}, permitting the exploration needed to discover how to satisfy them and allowing recovery from a violation rather than cutting the policy off at the first mistake.

The scale $c_{i}^{\max}$ makes the comparison in \eqref{eq:cat_delta} meaningful. Robotic constraints are heterogeneous in units and magnitude, so without normalization the maximum would be dominated by whichever constraint is expressed in the largest units. CaT therefore normalizes each violation by an exponential moving average of the largest violation that constraint produces in each batch of collected experience,
\begin{equation}
\label{eq:cat_norm}
c_{i}^{\max} \;\leftarrow\;
\tau_{c}\, c_{i}^{\max}
\;+\;
(1-\tau_{c}) \max_{(\bm{s},\bm{a})\in\mathrm{batch}} c_{i}^{+}(\bm{s},\bm{a}),
\end{equation}
with decay rate $\tau_{c}\in(0,1)$. The normalized quantity is dimensionless and lies in $[0,1]$ for all but the most extreme violations, so \eqref{eq:cat_delta} compares constraints on equal terms, and the adaptive update keeps the termination signal informative as violations shrink. This is what allows CaT to accept a large heterogeneous constraint set with no per-constraint tuning beyond $p_{i}^{\max}$, and is a principal reason for adopting it here. Note that \eqref{eq:cat_norm} maintains \emph{one} scalar scale per constraint, estimated from the largest violations that constraint produces \emph{anywhere in the task}; \cref{sec:method:cat_gating} examines the consequences of this in multi-phase settings.

Two further properties make CaT well suited to the present problem. Enforcement is \emph{state-wise}: $\delta$ is evaluated at every step from the instantaneous violation, so an impact transient is penalized at the impact and cannot be amortized against good behavior elsewhere in the episode. And the aggregation in \eqref{eq:cat_delta} is a maximum rather than a sum, so simultaneously active constraints do not compound into an overwhelming termination signal and further constraints can be added without retuning the others. In practice CaT requires only that rewards be scaled by $(1-\delta)$ and termination flags replaced by $\delta$, leaving the underlying algorithm unchanged.

%
%
%

\section{Method}
\label{sec:method}

\subsection{Overview}
\label{sec:method:overview}

The task defined in \cref{sec:problem} is solved by a single policy trained with model-free RL under a multi-stage extended version of the CaT formulation of \cref{sec:preliminaries}. The policy receives the noisy relative state of the target together with the servicer's proprioception, and emits incremental joint-position commands for both manipulators and a wrench command for the propulsion system. The joint commands are tracked by an underlying joint-space PD controller running at higher frequency, and the wrench command is subject to the actuator lag of \eqref{eq:actuator_lag}. No phase index is supplied, no controller is switched, and no grasp event is detected: the same network governs free-space closing, contact establishment, and detumbling.

\begin{figure*}[t]\centering
  \includegraphics[width=\linewidth]{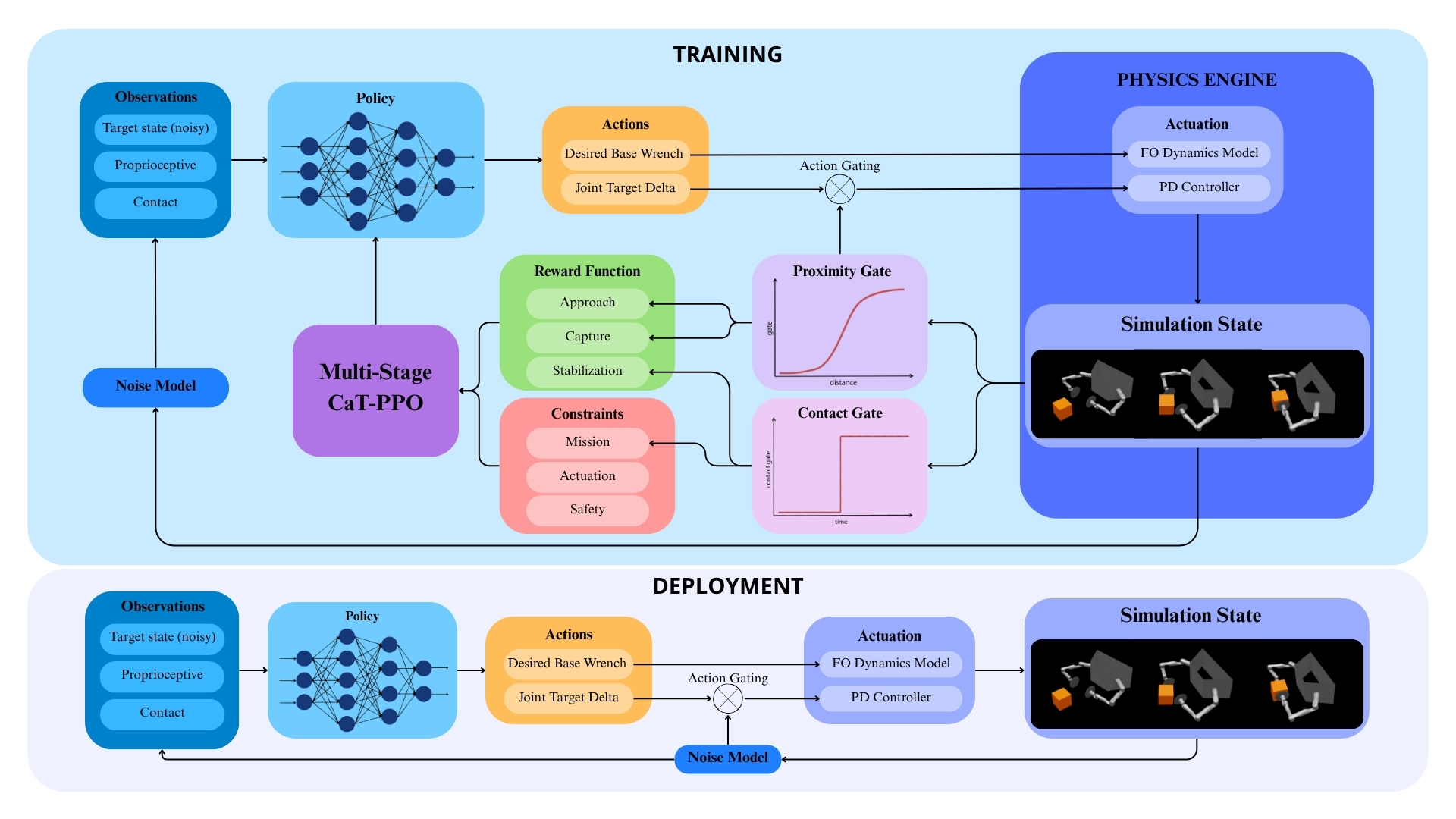}
  \caption{\textbf{Method overview}. Both gates are computed from the state, not
  from the policy: the proximity gate $g_{p}$ modulates the reward, the
  manipulator command and the constraint terms, while the contact gate $g_{c}$
  cascades onto the stabilization rewards. The reward and constraint machinery
  acts only during training; at deployment the phase structure it induced is
  already internal to the policy, which requires no supervision beyond the action
  gating of \cref{eq:action_gate}.}
  \label{fig:method_overview}
\end{figure*}

Both difficulties identified in \cref{sec:problem:statement}, namely the opposition between the approach and capture objectives, and the phase dependence of the constraint set, are resolved here by a single device. A scalar \emph{proximity gate} $g_{p}\in(0,1)$, a smooth function of the servicer--target separation, is applied consistently at three distinct loci of the CMDP (\cref{fig:method_overview}):
\begin{enumerate}
\item the \textbf{reward} (\cref{sec:method:reward}), where $g_{p}$ and its complement $1-g_{p}$ form a partition of unity that interpolates between the approach and capture objectives rather than summing them;
\item the \textbf{action space} (\cref{sec:method:action}), where $g_{p}$ multiplies the manipulator command, so that the arms are physically unable to act at range regardless of what the policy outputs;
\item the \textbf{constraint set} (\cref{sec:method:cat_gating}), where gating allows each constraint to be decomposed into phase-specific terms, each normalized on its own scale.
\end{enumerate}
A second gate $g_{c}$, derived from the quality of the contact rather than from proximity, cascades onto the stabilization objectives so that detumbling reward is reachable only through a well-formed grasp. The phase structure of the mission is thus never imposed; it is \emph{induced}, and the policy is free to discover where within the gate's transition band the behavioral change should occur.

\subsection{Observation and Action Spaces}
\label{sec:method:spaces}

\subsubsection{Observations}
The policy observes a $93$-dimensional vector comprising the noisy relative target state, the servicer's proprioception, the end-effector tracking errors, the measured contact forces, and the previous action, composed as given in \cref{tab:obs}.

\begin{table}[t]
\centering
\caption{\textbf{Observation space} ($93$ dimensions). Target quantities are corrupted by the estimation noise of \eqref{eq:noise}; the remainder are measured onboard. $\varphi(\cdot)$ denotes the $6$D rotation encoding \cite{zhou2019continuity}.}
\label{tab:obs}
\begin{tabular}{llcc}
\toprule
Group & Quantity & Symbol & Dim. \\
\midrule
\multirow{4}{*}{Target (noisy)}
  & Relative position, base frame  & $\tilde{\bm{p}}_{t}^{b}$          & 3 \\
  & Relative orientation           & $\varphi(\tilde{\bm{R}}_{t}^{b})$ & 6 \\
  & Relative linear velocity       & $\tilde{\bm{v}}_{t}^{b}$          & 3 \\
  & Angular velocity               & $\tilde{\bm{\omega}}_{t}^{b}$     & 3 \\
\midrule
\multirow{3}{*}{Servicer}
  & Base linear / angular velocity & $\bm{v}_{b}^{b},\,\bm{\omega}_{b}^{b}$ & 6 \\
  & Joint positions                & $\bm{q}$                               & 14 \\
  & Joint velocities               & $\dot{\bm{q}}$                         & 14 \\
\midrule
\multirow{2}{*}{End effectors}
  & Position error, both arms      & $\bm{e}_{p,\mathrm{L}},\,\bm{e}_{p,\mathrm{R}}$ & 6 \\
  & Orientation errors             & $\bm{e}_{R,\mathrm{L}},\,\bm{e}_{R,\mathrm{R}}$ & 12 \\
\midrule
Contact & Contact forces  & ${\bm{f}}_{\mathrm{L}},\,\bm{f}_{\mathrm{R}}$ & 6 \\
\midrule
History & Previous action          & $\bm{a}_{t-1}$ & 20 \\
\midrule
\textbf{Total} & & & \textbf{93} \\
\bottomrule
\end{tabular}
\end{table}

All orientations and orientation errors use the continuous $6$D representation of Zhou \emph{et al.}~\cite{zhou2019continuity}, which matters more here than in a fixed-base task: the target rotates continuously through $SO(3)$, and a discontinuous representation would present the network with an artificial jump in its input at an arbitrary attitude, in the middle of a contact-critical tracking task. This is also why the attitude noise of \eqref{eq:noise} is injected on $SO(3)$ before encoding: perturbing the six components directly would destroy their orthonormality, and
with it the continuity the representation is chosen for.

The end-effector errors are expressed relative to the commanded pose of \cref{sec:method:command}. For arm $i\in\{\mathrm{L},\mathrm{R}\}$ with current pose $(\bm{p}_{i}^{b},\bm{R}_{i}^{b})$ and command $(\bar{\bm{p}}_{i}^{b},\bar{\bm{R}}_{i}^{b})$,
\begin{equation}
\label{eq:ee_error}
\bm{e}_{p,i} = \bar{\bm{p}}_{i}^{b} - \bm{p}_{i}^{b},
\qquad
\bm{e}_{R,i} = \varphi\!\left( (\bm{R}_{i}^{b})^{\top}\bar{\bm{R}}_{i}^{b} \right).
\end{equation}


\subsubsection{Actions}
The policy emits $\bm{a}_{t}\in\mathbb{R}^{20}$, partitioned as
\begin{equation}
\label{eq:action_split}
\bm{a}_{t} =
\begin{bmatrix}
\bm{a}^{\mathrm{arm}}_{t} \in \mathbb{R}^{14} &
\bm{a}^{f}_{t} \in \mathbb{R}^{3} &
\bm{a}^{\tau}_{t} \in \mathbb{R}^{3}
\end{bmatrix}^{\!\top}\!\!,
\end{equation}
none of which is a physical quantity.

\paragraph{Manipulators}
The arm channel is an increment on the measured joint configuration, scaled and gated:
\begin{equation}
\label{eq:action_gate}
\bm{q}^{\mathrm{des}}_{t}
=
\bm{q}_{t} \;+\; g_{p}(d_{t})\; s_{q}\; \bm{a}^{\mathrm{arm}}_{t},
\qquad s_{q} = 0.01~\mathrm{rad},
\end{equation}
where $g_{p}$ is the proximity gate of \cref{sec:method:gate}, whose role here is the subject of \cref{sec:method:action}.


The policy thus regulates contact force through \eqref{eq:action_gate} and the underlying joint controller, without an explicit force-control loop. The single scale $s_{q}$ bounds the joint displacement per step, and therefore both how fast the arms slew and how far the setpoint is driven into the surface; the joint-rate bound itself is imposed by the actuator velocity limits of \cref{sec:problem:servicer}.

\paragraph{Propulsion}
The propulsion channels are clipped to $[-1,1]$ and scaled to the authority limits of \cref{sec:problem:servicer},
\begin{equation}
\label{eq:thruster_cmd}
\bm{f}^{\mathrm{cmd}}_{t} = f_{\mathrm{thr},\max}\,\bm{a}^{f}_{t},
\qquad
\bm{\tau}^{\mathrm{cmd}}_{t} = \tau_{\mathrm{rw},\max}\,\bm{a}^{\tau}_{t},
\end{equation}
with $f_{\mathrm{thr},\max}=10~\mathrm{N}$ and $\tau_{\mathrm{rw},\max}=5~\mathrm{N}\,\mathrm{m}$. The commanded wrench reaches the base only through the lag of \eqref{eq:actuator_lag}, and it is the delivered wrench $\bm{f}_{b}$ on which the actuation constraints of \cref{sec:method:constraints} are evaluated: the policy is constrained on what the thrusters \emph{deliver}, not on what it requests. The lag has a deliberate role. It denies the policy an instantaneous reversal of the base wrench, the maneuver it would otherwise learn in order to arrest
the approach at the last moment; it must instead anticipate the deceleration, and the base velocity constraints penalize it if it does not.

\subsubsection{End-effector command generation}
\label{sec:method:command}
The pose each end effector should attain is attached to a face of a tumbling
body, and therefore moves with it. At every control step, the desired pose of end effector $i\in\{\mathrm{L},\mathrm{R}\}$ is recomputed as the center of the corresponding target face, offset along the face normal by
\begin{equation}
\label{eq:press_offset}
d_{\mathrm{press}} = \tfrac{1}{2}\ell - \Delta,
\qquad \ell = 0.25~\mathrm{m},\;\; \Delta = 5~\mathrm{mm},
\end{equation}
so that the commanded pose lies a small distance \emph{inside} the face. The grasp is thereby specified as a press rather than a touch: the tracking error cannot be driven to zero without loading the surface, and the normal force required by \eqref{eq:friction_cone} emerges from pose tracking rather than having to be commanded separately. The desired \emph{orientation} co-rotates with the target, so the disk face remains parallel to the target face throughout the tumble. The manipulators must therefore track a moving goal, and capture becomes a continuous tracking problem that terminates in sustained contact.

\subsection{The Proximity Gate}
\label{sec:method:gate}

Let $d = \lVert \bm{p}_{t} - \bm{p}_{b}\rVert$ be the servicer--target separation. The proximity gate is
\begin{equation}
\label{eq:proximity_gate}
g_{p}(d) \;=\; \sigma\!\left(\frac{\bar{d} - d}{\kappa}\right),
\qquad
\sigma(x) = \frac{1}{1+e^{-x}},
\end{equation}
with threshold $\bar{d}$ and sharpness $\kappa$: it approaches $0$ far from the target, $1$ once the servicer has closed, and varies smoothly in between.

The calibration matters. The threshold is placed \emph{beyond} the stand-off: with $\bar{d} = d_{\star} + 0.10~\mathrm{m}$ and $d_{\star}\approx0.45~\mathrm{m}$, the gate reaches $g_{p}\approx0.97$ by the time the base arrives, so the capture objectives already dominate when the base settles, and the offset gives the manipulators a preview window in which to begin orienting toward the moving face. The sharpness $\kappa = 0.03~\mathrm{m}$ sets the transition width: narrow enough that the two regimes are distinct (the gate traverses $0.05$ to $0.95$ over roughly $0.18~\mathrm{m}$), yet wide enough that the objective changes over many control steps rather than discontinuously. A hard switch would reintroduce, inside the reward, the same handoff seam the method sets out to remove from the controller.

\subsection{Gating the Reward}
\label{sec:method:reward}

The reward is a weighted sum of terms, each gated by $g_{p}$, by its complement, or by neither, as listed in \cref{tab:rewards}. 

All shaping terms share a common form. Writing $\kappa(\bm{x};\sigma) =
\exp\!\left(-\lVert\bm{x}\rVert^{2}/\sigma^{2}\right)$ for a Gaussian kernel on
an error quantity $\bm{x}$ with width $\sigma$, every term of
\cref{tab:rewards} is $\kappa$ evaluated on the error it regulates, with the
exceptions noted in the table: the pose-tracking term uses the unsquared
argument $\exp(-\lVert\bm{x}\rVert/\sigma)$, summed over position and
orientation error, and the contact term is the centering-weighted saturation of
\eqref{eq:contact_reward}.

\begin{table}[t]
\centering
\caption{\textbf{Reward terms}. Each term is $\kappa(\bm{x};\sigma)$ on the
listed error unless marked; $g_{p}$ is the proximity gate
\eqref{eq:proximity_gate} and $g_{c}$ the contact gate \eqref{eq:contact_gate}.
All terms are scaled by the control timestep.}
\label{tab:rewards}
\setlength{\tabcolsep}{4pt}
\begin{tabular}{llccl}
\toprule
Term & Error $\bm{x}$ & $\sigma$ & Gate & $w$ \\
\midrule
\multicolumn{5}{l}{\emph{Approach}}\\
Closing distance & $d-d_{\star}$ & $0.25$~m & --- & 5.0 \\
Arm stillness & $\bm{q}-\bm{q}_{0}$ & $0.3$~rad & $1-g_{p}$ & 5.0 \\
\midrule
\multicolumn{5}{l}{\emph{Capture}}\\
EE pose tracking$^{\mathrm{a}}$ & $\bm{e}_{p}$; $\bm{e}_{R}$ & $0.3$; $0.5$ & $g_{p}$ & 10.0 \\
EE centering & $e_{i}$ & $0.10$~m & $g_{p}$ & 2.0 \\
Contact force$^{\mathrm{b}}$ & --- & --- & $g_{p}$ & 5.0 \\
\midrule
\multicolumn{5}{l}{\emph{Stabilization}}\\
Target linear vel. & $\bm{v}_{t}$ & $0.02$~m/s & $g_{c}g_{p}$ & 5.0 \\
Target angular vel. & $\bm{\omega}_{t}$ & $0.05$~rad/s & $g_{c}g_{p}$ & 5.0 \\
\midrule
\multicolumn{5}{l}{\emph{Base regulation}}\\
Base linear vel. & $\bm{v}_{b}$ & $0.1$~m/s & $g_{p}$ & 5.0 \\
Base angular vel. & $\bm{\omega}_{b}$ & $0.1$~rad/s & $g_{p}$ & 5.0 \\
\midrule
\multicolumn{5}{l}{\emph{Regularization}}\\
Arm action rate & $\Delta\bm{a}^{\mathrm{arm}}$ & $5.0$ & --- & 3.0 \\
Thruster action rate & $\Delta\bm{a}^{f}$ & $0.5$ & --- & 2.0 \\
Torque action rate & $\Delta\bm{a}^{\tau}$ & $0.5$ & --- & 2.0 \\
\bottomrule
\end{tabular}

\vspace{2pt}
\footnotesize
$^{\mathrm{a}}$ Unsquared argument, summed over the two errors.\quad
$^{\mathrm{b}}$ \eqref{eq:contact_reward}.
\end{table}

Three structural features carry the design.

\subsubsection{Complementary gating: a partition of unity}
The manipulator objective comprises two mutually exclusive behaviors, blended:
\begin{equation}
\label{eq:partition}
r_{\mathrm{arm}} \;=\;
\underbrace{\bigl(1-g_{p}\bigr)\, w_{\mathrm{s}}\, r_{\mathrm{stillness}}}_{\text{hold the home configuration}}
\;+\;
\underbrace{g_{p}\, w_{\mathrm{t}}\, r_{\mathrm{tracking}}}_{\text{track the co-rotating face}} .
\end{equation}
Because the two gates sum to unity, the total reward available to the arms is approximately constant in $d$; what changes is \emph{which behavior earns it}. The arm objective rotates continuously from one behavior to the other as the base closes, and at no point is the policy rewarded for doing both. A naive sum would reward tracking from any distance, including distances at which tracking is useless and, as \cref{sec:method:action} shows, actively harmful.

\subsubsection{Approach shaping}
The closing reward is a Gaussian on the \emph{signed} error $d - d_{\star}$, maximized \emph{at} the stand-off rather than at $d=0$. A monotone closing reward would encourage the servicer to drive into the target; a threshold reward would leave it free to loiter anywhere inside. Peaking at the stand-off penalizes both loitering and overshoot with a single term, and gives the base a well-defined station to hold once the manipulators take over.

\subsubsection{The contact gate: a two-level cascade}
A disk pressing on the corner of a tumbling cube transmits force but cannot sustain the tangential load that detumbling demands, and it can serve as a source of reward without progressing the mission. The stabilization objectives are therefore gated not on contact but on \emph{well-centered} contact. For each end effector, a centering weight is formed from the lateral offset $e_{i}$ between the contact point and the face center,
\begin{equation}
\label{eq:centering}
\chi_{i} \;=\; \exp\!\left(-\frac{e_{i}^{2}}{\sigma_{c}^{2}}\right),
\qquad \sigma_{c} = 0.10~\mathrm{m},
\end{equation}
and the contact reward weights the measured normal force by it,
\begin{equation}
\label{eq:contact_reward}
r_{\mathrm{contact}} \;=\;
g_{p}\sum_{i\in\{\mathrm{L},\mathrm{R}\}}
\tanh\!\left(\frac{f_{n,i}}{f_{\mathrm{sat}}}\right)\chi_{i},
\qquad f_{\mathrm{sat}} = 5~\mathrm{N}.
\end{equation}
The $\tanh$ saturates, so pressing harder than $f_{\mathrm{sat}}$ earns no further reward; together with the $20~\mathrm{N}$ contact-force constraint of \cref{sec:method:constraints}, this realizes the force corridor of \cref{sec:problem:contact}, rewarding the policy for pressing firmly while constraining it from crushing. The contact gate is derived directly from this quantity,
\begin{equation}
\label{eq:contact_gate}
g_{c} \;=\; \mathrm{clip}\!\left(\tfrac{1}{2}\, r_{\mathrm{contact}},\; 0,\; 1\right),
\end{equation}
and the stabilization rewards are gated by the \emph{product} $g_{c}\,g_{p}$. 

The result is a cascade rather than a flat gate: reward for detumbling is accessible only if the contact is centered, which is accessible only if the arms are tracking, which is accessible only if the base has closed. The policy cannot collect stabilization reward by nudging the target with an arm tip, nor before it has earned a grasp. Each phase's reward is unlocked by completion of the previous one, without any phase index being computed or any transition declared.

\subsection{Gating the Action Space}
\label{sec:method:action}

Gating the reward expresses a preference; it does not remove a capability. The second locus does. In \eqref{eq:action_gate} the manipulator command is scaled by the same gate, so that as $g_{p}\rightarrow 0$ the commanded displacement vanishes and the arms hold their configuration whatever the policy emits: far from the target, the manipulator sub-space of the action is in effect disabled.

The reward gating does not make this redundant, for a reason specific to free-flying manipulation. The base masses $150~\mathrm{kg}$ and the target as little as $15~\mathrm{kg}$, so a policy that has learned only that reward accrues as $d$ decreases has a second solution available through the momentum coupling of \eqref{eq:eom}, and it is cheaper than thrusting the base: extend
an arm, contact the target, and pull. The target accelerates inward, $d$ decreases, the closing reward is collected, and meanwhile the debris has been given an uncontrolled velocity, no grasp has formed, and the base has been disturbed by the reaction. 

Reward shaping can penalize this exploit but cannot forbid it, and a randomly exploring policy may find it first and never leave it. Equation~\eqref{eq:action_gate} removes it from the action space instead: while $d$ is large the arms cannot reach out at all, so the only way to close is to move the base, and manipulator authority appears only once $g_{p}$ has risen, by which point the reward of \cref{sec:method:reward} rewards the arms for tracking the face rather than for reducing the separation. It is essential that both loci are driven by the \emph{same} signal: sharing $g_{p}$ guarantees the manipulators acquire authority no faster than they acquire the objective it serves. \cref{sec:results} reports an ablation isolating this locus.

\subsection{Gating the Constraints}
\label{sec:method:cat_gating}

The third locus concerns the constraint set, and addresses a difficulty that is a property of CaT itself rather than of the ADR task.

\subsubsection{The normalization problem}
The normalization of \eqref{eq:cat_norm} carries an implication that is benign in the single-regime tasks for which CaT was developed, and damaging here. Because one scalar scale is estimated from the largest violations a constraint produces \emph{anywhere} in the task, the pressure that constraint exerts in a given situation depends not on the violation there alone, but on that violation \emph{relative to the worst it incurs anywhere}. Consider a base-velocity constraint imposed as a single always-active term at the tight limit capture requires, $0.01~\mathrm{m}\,\mathrm{s}^{-1}$. The approach translates the base at $\sim\!0.1~\mathrm{m}\,\mathrm{s}^{-1}$, a violation of $\sim\!0.09~\mathrm{m}\,\mathrm{s}^{-1}$ that enters every batch, so \eqref{eq:cat_norm} settles at $c^{\max}\approx0.09$. During capture the residual drift threatening the grasp is a violation of only $\sim\!0.01~\mathrm{m}\,\mathrm{s}^{-1}$; normalized, its magnitude is $0.01/0.09\approx0.11$, and with $p^{\max}=0.25$ the induced termination probability is

\begin{equation}
\label{eq:dilution}
\delta \;=\; 0.25 \times 0.11 \;\approx\; 0.03 .
\end{equation}
The constraint that exists to enforce a reactionless base exerts essentially no pressure during capture: its normalization has been captured by the approach, and against those transients the drift that actually threatens the grasp appears negligible. It is not badly tuned; it has been diluted. This follows directly from \eqref{eq:cat_norm} and arises in any multi-phase task whose phases load the same constraint at different magnitudes.

\subsubsection{Phase-gated constraint decomposition}
The remedy follows from the diagnosis: if one scale is being asked to serve two regimes, the constraint should be split so that each regime carries its own. Each base-velocity constraint is therefore replaced by a \emph{pair},
\begin{align}
\label{eq:decomposition}
c^{\mathrm{loose}}(\bm{s},\bm{a}) &= \lVert \bm{v}_{b}\rVert - v^{\mathrm{loose}},
\\[2pt]
c^{\mathrm{tight}}(\bm{s},\bm{a}) &= \bigl(\lVert \bm{v}_{b}\rVert - v^{\mathrm{tight}}\bigr)\cdot \mathbb{1}_{c},
\end{align}
where $v^{\mathrm{loose}} \gg v^{\mathrm{tight}}$ and $\mathbb{1}_{c}$ indicates end-effector contact. The loose term is always active and shapes the approach, capping the closing speed and forcing the servicer to decelerate rather than arrive ballistically. The tight term is identically zero out of contact, and therefore contributes nothing to its own batch maximum during the approach: its scale is estimated exclusively from contact-phase violations. That term now sees its $0.01~\mathrm{m}\,\mathrm{s}^{-1}$ violation against a scale of the same order, a normalized magnitude near unity, so with $p^{\max}=0.5$
\begin{equation}
\label{eq:decomposed_delta}
\delta \;\approx\; 0.5 \times 1.0 \;=\; 0.5 ,
\end{equation}
more than an order of magnitude above \eqref{eq:dilution} on the same physical violation. The tight limit is enforced where it matters; the loose limit continues to shape the phase where the tight one would be nonsensical.

The same decomposition is applied to the target-velocity constraints, where the motivation is complementary. The target tumbles freely during the approach at rates far above any limit the mission would tolerate after capture, so an always-active constraint on target angular velocity would be violated massively throughout the approach, for a condition the servicer has no physical means of influencing and, through the maximum in \eqref{eq:cat_delta}, would suppress the entire reward signal from the first step of every episode. Gating on contact restricts it to the regime in which the policy can act on it.

The constraint gates are binary indicators, not the smooth sigmoids of \cref{sec:method:gate}, and the distinction is deliberate: the reward and action gates shape a continuous landscape differentiated through by the learning signal, whereas the constraint gates feed only the violation magnitude to \eqref{eq:cat_delta}, with no gradient taken through them. A binary indicator is therefore sufficient, and makes the tight term \emph{exactly} zero out of contact, which is what keeps its normalization scale uncontaminated. A smooth gate would leak a small violation at every approach step, and that leakage, multiplied by the large approach-phase excursion, would reintroduce the contamination the decomposition exists to prevent.
\subsection{Constraint Set}
\label{sec:method:constraints}

\Cref{tab:constraints} lists the full constraint set. Constraints fall into three
groups:

\emph{Mission constraints} bound the residual motion of the target after capture
and the reaction transmitted to the base during pressing. These are the
decomposed pairs of \cref{sec:method:cat_gating}.

\emph{Actuation constraints} bound thruster force, reaction torque, and joint
effort. These are set \emph{below} the corresponding hardware limits: the joint-torque constraint is $5~\mathrm{N}\,\mathrm{m}$ against actuators capable of $39$ and $13~\mathrm{N}\,\mathrm{m}$, and the base force constraint is $5~\mathrm{N}$ against $10~\mathrm{N}$ of thruster
authority. They do not encode what the hardware can survive; they encode the
margin within which the mission is prepared to operate, and they leave the
remaining authority as reserve.

\emph{Safety constraints} bound the contact force at the end effectors and
prohibit contact between the target and the servicer base. The contact-force
limit of $20~\mathrm{N}$ is the upper bound of the corridor established in \cref{sec:problem:contact}. Base collision is the sole hard
constraint, with $p^{\max}=1$: a collision between the debris and the servicer
body is the failure the mission exists to avoid, and it also terminates the
episode.

\begin{table}[t]
\centering
\caption{\textbf{Constraint set}. $\mathbb{1}_{c}$ denotes the end-effector contact
indicator. $p^{\max}$ is the maximum termination probability of
\eqref{eq:cat_delta}.}
\label{tab:constraints}
\begin{tabular}{llcc}
\toprule
Constraint & Limit & Gate & $p^{\max}$ \\
\midrule
\multicolumn{4}{l}{\emph{Mission}}\\
Target linear velocity   & $0.01~\mathrm{m/s}$    & $\mathbb{1}_{c}$ & 0.25 \\
Target angular velocity  & $0.02~\mathrm{rad/s}$  & $\mathbb{1}_{c}$ & 0.25 \\
Base linear velocity (loose)  & $0.08~\mathrm{m/s}$   & ---              & 0.25 \\
Base angular velocity (loose) & $0.10~\mathrm{rad/s}$ & ---              & 0.25 \\
Base linear velocity (tight)  & $0.01~\mathrm{m/s}$   & $\mathbb{1}_{c}$ & 0.50 \\
Base angular velocity (tight) & $0.01~\mathrm{rad/s}$ & $\mathbb{1}_{c}$ & 0.50 \\
\midrule
\multicolumn{4}{l}{\emph{Actuation}}\\
Thruster force (per axis)  & $5~\mathrm{N}$            & --- & 0.60 \\
Reaction torque (per axis) & $2~\mathrm{N\,m}$         & --- & 0.60 \\
Joint torque (per joint)   & $5~\mathrm{N\,m}$         & --- & 0.25 \\
\midrule
\multicolumn{4}{l}{\emph{Safety}}\\
End-effector contact force & $20~\mathrm{N}$           & --- & 0.90 \\
Base collision             & $1~\mathrm{N}$            & --- & \textbf{1.00} \\
\bottomrule
\end{tabular}
\end{table}


\subsection{Curriculum}
\label{sec:method:curriculum}

Training begins from a task the policy can solve and expands toward the task the
mission requires. A single scalar $\lambda\in[0,1]$, advancing linearly with
training and saturating at $70\%$ of the schedule, drives two ranges.

\emph{Tumbling rate:} The initial angular velocity of the target is sampled from
a range that grows with $\lambda$, from a stationary target at $\lambda=0$ to the
full rate of \cref{sec:problem:target} at $\lambda=1$.

\emph{Initial separation:} The initial distance is sampled from
$[d_{\star},\, d_{\star} + 0.5\lambda~\mathrm{m}]$. The range \emph{expands};
it does not translate. Near-contact initial conditions are retained at every
stage of training, so the policy continues to rehearse capture and detumbling
while it is learning to approach, and does not forget the contact skills while
acquiring the free-space ones.

The expanding-range design has a consequence worth making explicit, because it
frames what the approach phase contributes. At $\lambda = 0$ the servicer begins
already on station against a stationary target: the task degenerates to the capture-and-detumble problem, with no approach to perform. The curriculum therefore teaches the contact-intensive task first, and introduces the approach phase smoothly afterwards.

\subsection{Training}
\label{sec:method:training}

The policy is trained with a customized implementation of
PPO~\cite{schulman2017ppo} under the CaT objective (\cref{eq:cat_objective}), extended by the machinery in \cref{sec:method:cat_gating} to enable multi-stage tasks,
using $4096$ environments simulated in parallel on a single
GPU~\cite{makoviychuk2021isaacgym, mittal2025isaaclab, rudin2022walk}. Physics is
integrated at $200~\mathrm{Hz}$ and the policy queried at $50~\mathrm{Hz}$;
episodes last $T=30~\mathrm{s}$. The target mass is randomized over the range of
\cref{sec:problem:target} at every reset, and the relative-state observations are
corrupted according to \cref{eq:noise}.

The actor and critic are multilayer perceptrons with three hidden layers of
[256,256,256] units and tanh activations. Optimization uses
Adam at a learning rate of $0.001$, with discount factor $\gamma=0.995$, GAE
parameter $\lambda=0.95$, clip range 0.2, and entropy coefficient
0.005; 24 steps are collected per environment per update, over
2500 epochs and 4 mini-batches. The constraint normalization of
\cref{eq:cat_norm} uses a decay rate $\tau_{c}=0.95$. Training takes approximately 50 minutes on a single desktop machine with an NVIDIA GeForce 3090Ti GPU.

\section{Experiments and Results}
\label{sec:results}

The evaluation is organized around the contributions of \cref{sec:intro}.
\Cref{sec:results:wholetask} establishes that the single policy solves the
complete mission and quantifies its performance over a large randomized trial
set; \cref{sec:results:constraints} reports its constraint satisfaction; and
\cref{sec:results:failure} analyses the residual failures.
\Cref{sec:results:emergent} examines the phase structure the policy acquires
without supervision. \Cref{sec:results:generalization} then probes how far that
competence extends beyond the distribution the policy was trained on, across
target geometry, inertial asymmetry, tumbling rate and mass. Finally, \cref{sec:results:ablations} isolates the design choices themselves, and
establishes that the gating of \cref{sec:method:reward,sec:method:action} and the
constraint decomposition of \cref{sec:method:cat_gating} are necessary rather than
incidental, and \cref{sec:results:comparison} situates the method against
representative model-based and hybrid controllers.

Unless stated otherwise, all results are obtained from a single trained policy,
evaluated with the observation noise of \eqref{eq:noise} active, on trials whose
target mass, initial separation, and initial tumbling rate are sampled
independently across the full ranges of \cref{sec:problem}. No phase
supervision, no target inertial parameters, and no ground-truth target state are
provided to the policy at any point during evaluation.

\subsection{Evaluation Protocol and Metrics}
\label{sec:results:protocol}

The policy is evaluated over $N=1000$ independent episodes, each initialized
with a target mass drawn uniformly from $[15,45]~\mathrm{kg}$ and an initial
separation drawn uniformly from $[0.45, 0.95]~\mathrm{m}$, the approach range of
\cref{sec:method:curriculum}. Each component of the initial angular velocity is
drawn uniformly from $\pm0.25~\mathrm{rad}\,\mathrm{s}^{-1}$, resulting in a maximum magnitude of $0.25\sqrt{3} \approx 0.43~\mathrm{rad}\,\mathrm{s}^{-1}$. Every
episode runs to the $30~\mathrm{s}$ horizon; no episode is reset early except on
the collision termination of \cref{sec:method:constraints}.

An episode is scored a \emph{success} when the full condition of
\eqref{eq:success} holds continuously over the final second of the episode: the
target linear and angular speeds remain simultaneously below $\varepsilon_{v}$
and $\varepsilon_{\omega}$ while both end effectors remain in contact. Requiring the condition to be \emph{held} over a dwell window, rather than merely met at the final instant, rejects trials that cross the tolerance only transiently; requiring the grasp to be maintained rejects a policy that arrests the tumble but lets the target drift or loses the friction grasp.

Alongside the binary success rate, three families of metrics characterize the
outcome. \emph{Phase completion rates} decompose the mission into the fractions
of episodes that complete the approach, establish a bilateral grasp, and reduce
the target to rest, and so localize where any failures occur. \emph{Residual
metrics} report the final target and base velocities, the timing of
stabilization, the contact forces, and the total $\Delta v$ expended.
\emph{Constraint violation rates} report, per constraint, the fraction of
control steps at which its limit is exceeded. Reported intervals on rates are
$95\%$ Wilson score intervals.
\subsection{Whole-Task Performance}
\label{sec:results:wholetask}
 
\Cref{tab:expA_funnel} reports the phase completion rates over the $1000$ trials.
The policy completes the terminal approach and establishes a bilateral friction
grasp in every trial, and reduces the target to rest in $99.5\%$ of them
($95\%$ CI $[98.8, 99.8]$). Because the approach and grasp stages complete
without exception, the whole-task success rate is set entirely by the
stabilization stage: the single policy fails to bring five of the thousand targets fully to rest
within the horizon, and succeeds on the remainder. This is the primary result of
the paper.
 
\Cref{tab:expA_residuals} reports the residual state and effort over the
successful trials. The target is brought to a residual angular speed of
$0.0036 \pm 0.0024~\mathrm{rad}\,\mathrm{s}^{-1}$ and a residual linear speed of
$0.0020 \pm 0.0009~\mathrm{m}\,\mathrm{s}^{-1}$, in both cases well inside the
success tolerances. These residuals are reached from a mean initial tumbling rate
of $0.24~\mathrm{rad}\,\mathrm{s}^{-1}$, so the policy removes more than $98\%$ of the initial angular velocity while maintaining the grasp.
 
Two features of \cref{tab:expA_residuals} bear on claims made earlier in the
paper. First, the residual velocities of the servicer base ($0.0028~\mathrm{m}\,\mathrm{s}^{-1}$ linear and
$0.0040~\mathrm{rad}\,\mathrm{s}^{-1}$ angular) are of the same order as
those of the detumbled target, confirming that the base is held effectively
reactionless throughout the pressing and detumbling phases rather than being
pushed off station by the contact wrenches it applies. Second, the contact forces occupy the corridor that the reward and constraint design of \cref{sec:method:reward,sec:method:constraints} was intended to produce. The steady-state normal force settles at $11.0 \pm 0.3~\mathrm{N}$, comfortably above the $5~\mathrm{N}$ at which the contact reward saturates and comfortably
below the $20~\mathrm{N}$ constraint limit, with peak forces of
$14.6 \pm 4.3~\mathrm{N}$ during the initial press transient: firm enough to sustain the friction grasp against an unknown inertia, without approaching the ceiling that would risk damaging the target. The mission is completed at a mean cost of $0.045~\mathrm{m}\,\mathrm{s}^{-1}$ of
$\Delta v$.

\begin{table}[t]
\centering
\caption{\textbf{Phase completion rates over $n=1000$ randomized trials}. Intervals are
$95\%$ Wilson score intervals.}
\label{tab:expA_funnel}
\begin{tabular}{lc}
\toprule
Stage & Rate [95\% CI] (\%) \\
\midrule
Approach complete & 100.0 [99.6, 100.0] \\
Grasp established & 100.0 [99.6, 100.0] \\
Detumble complete (kinematic) & 99.5 [98.8, 99.8] \\
\textbf{Full success (grasp held)} & \textbf{99.5 [98.8, 99.8]} \\
\bottomrule
\end{tabular}
\end{table}
 
\begin{table}[t]
\centering
\caption{\textbf{Residual state, timing, and actuation effort at episode end}, mean $\pm$
std over the $n=995$ successful trials.}
\label{tab:expA_residuals}
\begin{tabular}{ll}
\toprule
Quantity & Value \\
\midrule
Residual target linear velocity $\lVert\bm{v}_t\rVert$   & $0.0020 \pm 0.0009$~m/s \\
Residual target angular velocity $\lVert\bm{\omega}_t\rVert$ & $0.0036 \pm 0.0024$~rad/s \\
Residual base linear velocity $\lVert\bm{v}_b\rVert$     & $0.0028 \pm 0.0012$~m/s \\
Residual base angular velocity $\lVert\bm{\omega}_b\rVert$ & $0.0040 \pm 0.0014$~rad/s \\
\midrule
Time to full stabilization (mean)   & $9.7 \pm 8.9$~s \\
Time to full stabilization (median) & $5.9$~s \\
\midrule
Peak contact force (per EE)         & $14.6 \pm 4.3$~N \\
Steady-state contact force (per EE) & $11.0 \pm 0.3$~N \\
\midrule
Total $\Delta v$ expenditure        & $0.045 \pm 0.012$~m/s \\
\bottomrule
\end{tabular}
\end{table}
 
\begin{figure}[t]\centering
  \includegraphics[width=\linewidth]{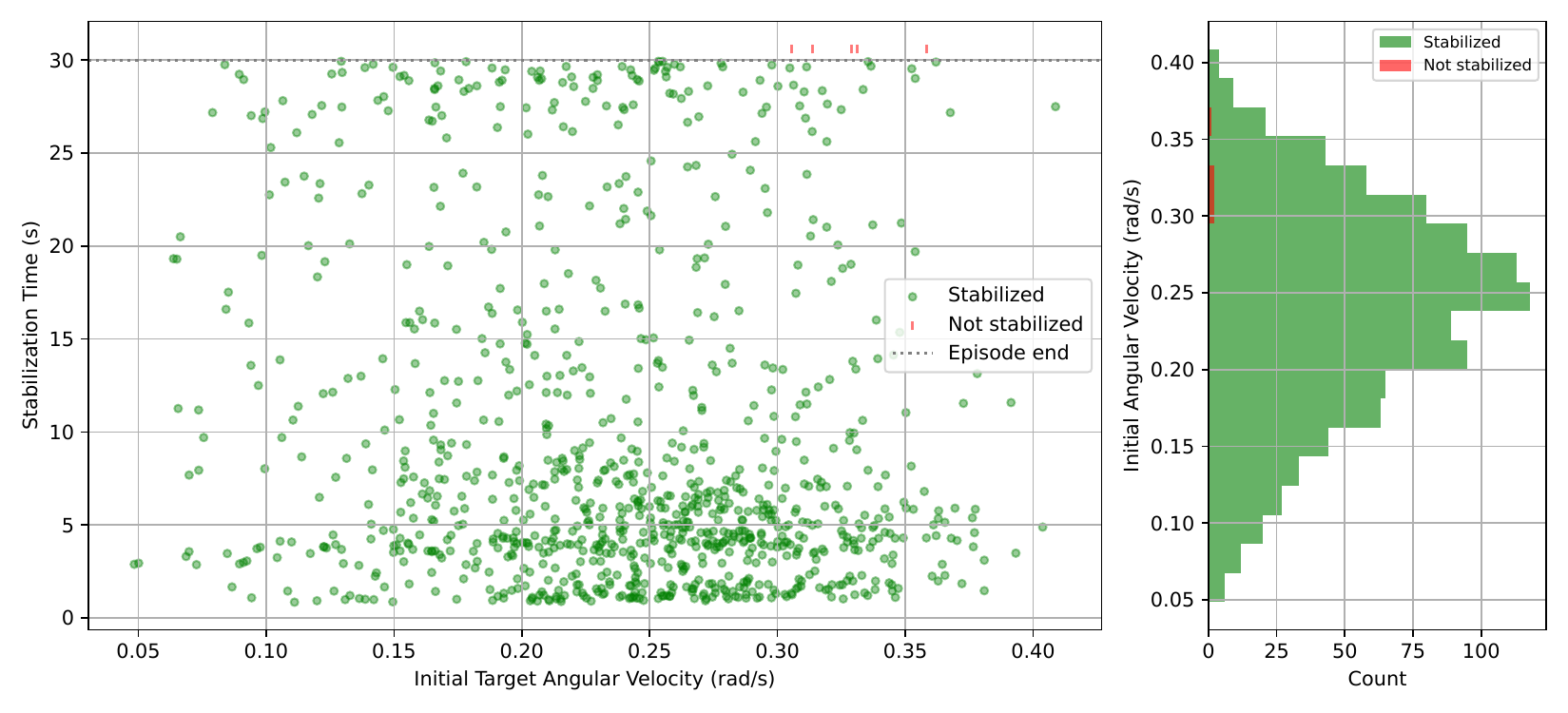}
  \caption{\textbf{Time to full stabilization against initial target angular velocity
  over $1000$ trials}; the five failures are marked at the episode horizon. The
  side panel shows the distribution of the initial tumbling magnitude.}
  \label{fig:expA_stabtime}
\end{figure}

The time to stabilization is the one metric whose distribution is strongly
asymmetric: its median of $5.9~\mathrm{s}$ lies well below its mean of
$9.7~\mathrm{s}$, so most targets are brought to rest quickly while a minority
extend the tail toward the horizon. \Cref{fig:expA_stabtime} plots the stabilization time against the initial tumbling rate, and the two are largely independent: the spread at any given rate exceeds the variation across rates. The time required is therefore governed less by the momentum to be removed
than by the precision to which the residual must be driven, a reading
reinforced in \cref{sec:results:gen:mass}, where a tenfold increase in target
inertia \emph{reduces} the settling time rather than extending it. The five
failures are the only trials to reach the horizon without satisfying the success
condition, and all five lie in the upper part of the tumbling-rate range; they
are examined in \cref{sec:results:failure}.
 
How these quantities evolve within an episode is not resolved by aggregate statistics; \cref{sec:results:emergent} examines it on a representative trial.

\subsection{Constraint Satisfaction}
\label{sec:results:constraints}

\Cref{tab:expA_constraints} reports the fraction of control steps at which each
constraint of \cref{tab:constraints} is violated, over the same $1000$ trials.

The \emph{safety} constraints are satisfied at all times to within a fraction of a
percent. The base-collision constraint is never violated across the full trial
set, and the end-effector contact-force constraint that protects the target from
being crushed is exceeded in only $0.27\%$ of steps, during the brief press
transient at first contact. The \emph{actuation} constraints on thruster force,
reaction torque, and joint torque are each violated in no more than $0.12\%$ of steps,
confirming that the policy operates within the mission-level margins imposed in
\cref{sec:method:constraints} rather than at the limits of the hardware.

The \emph{mission} constraints register higher rates, and this is by design. The
target-velocity constraints are contact-gated: they become active once contact is
established and apply termination pressure toward rest while the target is still
being detumbled, so a rate of a few percent reflects the fraction of the episode
spent actively arresting the tumble rather than any failure of enforcement. The
decomposed base-velocity constraints make the effect of the design of
\cref{sec:method:cat_gating} directly visible in the data: the always-active
loose terms, which shape the approach, are essentially never violated
($0.00\%$ and $0.01\%$), while their contact-gated tight partners register the
residual base motion during pressing ($1.43\%$ and $2.99\%$). Were these enforced
as single constraints with one normalization scale, the tight limits would be
diluted by the approach-phase transient and would exert little pressure during
capture; the decomposition is what keeps them effective in the phase where they
matter, as \cref{sec:results:ablations} examines directly.

The temporal structure behind these rates is visible in
\cref{fig:emergent_phase}, where the loose base-velocity limits bound the
approach and the tight limits engage at contact, the residual base motion
remaining below them for the remainder of the episode.

\begin{table}[t]
\centering
\caption{\textbf{Constraint satisfaction over the $1000$ trials of
\cref{sec:results:wholetask}}, as the percentage of control steps at which each limit is exceeded.}
\label{tab:expA_constraints}
\begin{tabular}{llc}
\toprule
Constraint & Symbol & Violation (\%) \\
\midrule
\multicolumn{3}{l}{\emph{Mission}}\\
Target linear velocity & $c_{v_t}$ & 2.25 \\
Target angular velocity & $c_{\omega_t}$ & 3.55 \\
Base linear velocity (loose) & $c_{v_b}$ & 0.00 \\
Base angular velocity (loose) & $c_{\omega_b}$ & 0.01 \\
Base linear velocity (tight) & $c_{v_b}^{\,c}$ & 1.43 \\
Base angular velocity (tight) & $c_{\omega_b}^{\,c}$ & 2.99 \\
\midrule
\multicolumn{3}{l}{\emph{Actuation}}\\
Thruster force (per axis) & $c_{f}$ & 0.12 \\
Reaction torque (per axis) & $c_{\tau}$ & 0.06 \\
Joint torque (per joint) & $c_{q}$ & 0.02 \\
\midrule
\multicolumn{3}{l}{\emph{Safety}}\\
End-effector contact force & $c_{F}$ & 0.27 \\
Base collision & $c_{\mathrm{col}}$ & \textbf{0.00} \\
\bottomrule
\end{tabular}
\end{table} 

\subsection{Failure Analysis}
\label{sec:results:failure}

Five of the $1000$ trials did not meet the success condition, and in every one the policy captures the target and substantially detumbles it (\cref{tab:expA_failures}). Each completes the approach, establishes a bilateral grasp and holds it to the end, satisfies the linear-velocity tolerance, and leaves the base at rest; none involves a safety-constraint violation. Only the target angular velocity remains outside specification, and only after the tumble has been almost entirely removed: all five reach between $0.0008$ and $0.0041~\mathrm{rad}\,\mathrm{s}^{-1}$ at some point, more than an order of magnitude inside the tolerance. These are targets at the upper end of the difficulty range, with a mean initial tumbling rate of $0.33~\mathrm{rad}\,\mathrm{s}^{-1}$ against a population mean of $0.24$, and the persistence of the residual is not a consequence of the horizon: the minimum is reached between $t=2.9$ and $t=28.9~\mathrm{s}$, leaving a substantial part of most episodes after the target has first been brought to rest.

\begin{table}[t]
\centering
\caption{\textbf{The five failed trials of \cref{sec:results:wholetask}}.
$\min\lVert\bm{\omega}_t\rVert$ is the lowest angular velocity reached, at $t_{\min}$; $\lVert\bm{\omega}_t\rVert_f$ is the value at the end of the episode. The tolerance is $\varepsilon_\omega=0.03$~rad/s.}
\label{tab:expA_failures}
\begin{tabular}{rccccc}
\toprule
Sample ID & $\lVert\bm{\omega}_t\rVert_0$ & $\min\lVert\bm{\omega}_t\rVert$ & $t_{\min}$ & $\lVert\bm{\omega}_t\rVert_f$ & $m_t$ \\
    & (rad/s) & (rad/s) & (s) & (rad/s) & (kg) \\
\midrule
38  & 0.329 & 0.0041 & 28.9 & 0.0289 & 30.9 \\
550 & 0.314 & 0.0016 & 12.7 & 0.0584 & 33.4 \\
578 & 0.358 & 0.0018 &  8.2 & 0.0192 & 40.5 \\
662 & 0.305 & 0.0008 &  2.9 & 0.0259 & 44.4 \\
825 & 0.331 & 0.0009 &  6.4 & 0.0445 & 26.6 \\
\bottomrule
\end{tabular}
\end{table}

The five divide into two groups. In three (samples $38$, $578$ and $662$) the final angular velocity lies \emph{below} the tolerance, at $0.019$ to $0.029~\mathrm{rad}\,\mathrm{s}^{-1}$: these fail on the dwell requirement rather than the tolerance, since \eqref{eq:success} must hold at every step of the final second and the residual necessarily crosses the limit somewhere within the window. Under an instantaneous end-of-episode criterion all three would have been scored successful. In the remaining two, the target is left at $0.058$ and $0.045~\mathrm{rad}\,\mathrm{s}^{-1}$ and the trials fail on the specification itself, though both had earlier been brought below $0.002~\mathrm{rad}\,\mathrm{s}^{-1}$. In every case the outcome is benign: the grasp is retained, the servicer is stable, and the target is left rotating slowly rather than released.


\subsection{Emergent Phase Structure}
\label{sec:results:emergent}

The policy receives no phase index, evaluates no switching condition, and selects no controller: a single network governs the episode throughout. This raises a fair question: whether a policy modulated by a distance-dependent gate has merely had a hand-designed switch substituted for a learned one.

It has not, and the distinction carries the contribution. The gates determine
\emph{where}, as a function of state, reward and manipulator authority are
placed; the actions taken within that placement remain the policy's own. Nothing in the proximity gate instructs the servicer to thrust toward the target, to arrest that motion at the stand-off rather than overshoot, to time the arms onto a rotating face, or to modulate two friction forces so as to detumble an unknown inertia while canceling its own base reaction. Each is a control problem the policy solves, not a branch it is switched through. 

\begin{figure*}[t]\centering
  \includegraphics[width=\linewidth]{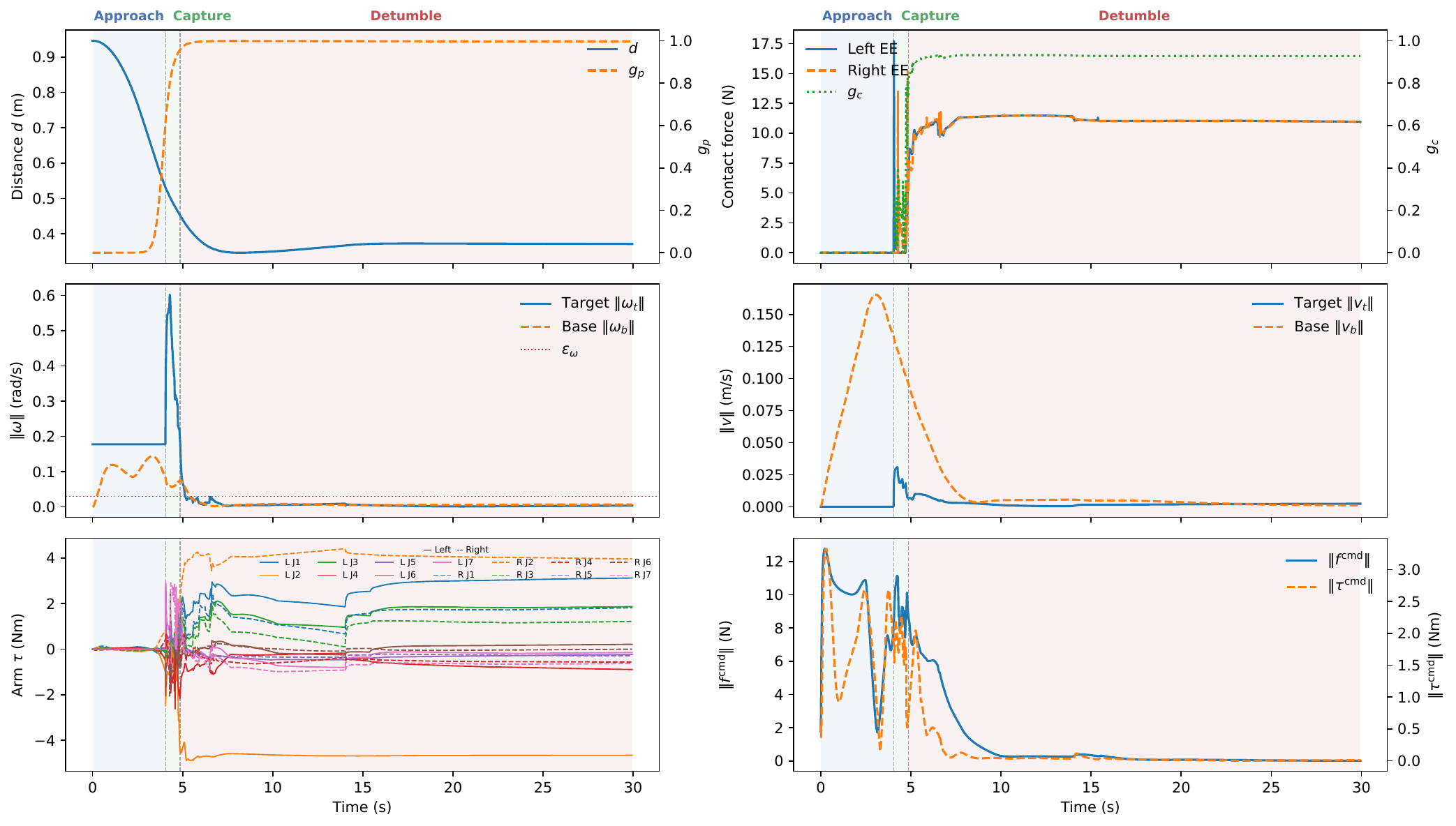}
  \caption{\textbf{Emergent phase structure in one representative episode}. Left column, top to bottom: servicer--target distance with the proximity gate $g_p$; target and base angular velocity with the success tolerance $\varepsilon_\omega$; manipulator joint torques (solid: left arm;
  dashed: right arm). Right column, top to bottom: end-effector contact forces with the contact gate $g_c$; target and base linear velocity; commanded thruster force and reaction torque magnitudes. Vertical lines mark the behaviorally-defined phase boundaries --- the onset of bilateral contact and the onset of sustained tumble reduction --- rather than any gate value.}
  \label{fig:emergent_phase}
\end{figure*}

\Cref{fig:emergent_phase} traces one representative episode: the successful trial whose initial tumbling rate and mass are closest to the population medians and whose initial separation lies in the upper quartile, so that a full approach is exercised. The regime boundaries drawn across the panels are defined not by the gates but by physical events: the onset of bilateral contact, and the onset of sustained tumble reduction. The accompanying video shows the same episode in motion, together with representative trials from the generalization scenarios of
\cref{sec:results:generalization}.

The approach appears in the distance, propulsion and joint-torque panels together. The servicer applies a burst of thrust, closes the separation, then cuts the command well before arrival, coasting onto the stand-off rather than braking against it. Nothing in the formulation requires this: the gate is a function of distance alone and issues no deceleration command, the closing reward peaks \emph{at} the stand-off rather than at contact, and the actuator lag of \eqref{eq:actuator_lag} denies an instantaneous reversal. Meanwhile the joint torques of both arms remain essentially constant, holding the home configuration. That guarantee is structural rather than learned: \cref{eq:action_gate} enforces it, the reward does not, and it forecloses the grab-and-pull behavior of \cref{sec:method:action}.

Contact establishment follows in the force panel: both end-effector forces rise to their regulated steady value within a fraction of a second and hold there, the contact gate $g_c$ rising with them. It reads out the centered bilateral pressure the policy has achieved; it does not produce it. The joint torques rise at the same instant and settle to the values that maintain the press, while the target linear velocity shows the small perturbation imparted by contact and its regulation back toward zero.

The angular-velocity panel carries the central observation. The target angular velocity falls to within the tolerance $\varepsilon_\omega$, while the base angular velocity remains near zero throughout and shows \emph{no step} at the onset of contact: the transition from free-space approach to contact-rich capture imparts no impulsive disturbance to the servicer. A modular controller that estimated the phase and switched control laws at contact would, at best, drive this quantity to zero on either side of a discontinuity at the switch; the single policy never incurs the discontinuity, because there is no switch. The ``open seam'' at the approach--capture boundary, identified in \cref{sec:intro} as the point at which model-based pipelines are most fragile,
does not arise. The magnitudes evolve smoothly throughout: the separation closes without overshoot, the thrust command is applied and withdrawn without chattering, the contact forces rise and hold, the joint torques settle rather than oscillate, and both target velocities decay without excursion.

The panels also allow the constraint satisfaction of \cref{sec:results:constraints} to be read directly. Before contact the base velocities are bounded by the loose limits of $0.08~\mathrm{m}\,\mathrm{s}^{-1}$ and $0.10~\mathrm{rad}\,\mathrm{s}^{-1}$, which admit the closing maneuver; once contact is established the tight limits of $0.01~\mathrm{m}\,\mathrm{s}^{-1}$ and $0.01~\mathrm{rad}\,\mathrm{s}^{-1}$ come into force and both fall to the order of $10^{-3}$ for the remainder of the episode. This is the decomposition of \cref{sec:method:cat_gating} operating as intended: a single limit could not both permit the approach and hold the base to this tolerance during capture. The joint torques stay within the $5~\mathrm{N\,m}$ per-joint limit throughout, including across the press transient where the largest excursions occur.


\subsection{Generalization and Robustness}
\label{sec:results:generalization}
 
The evaluation of \cref{sec:results:wholetask} samples the distribution the
policy was trained on. This subsection asks how far that competence extends
beyond it, along two distinct axes. The first is \emph{transfer to conditions
never encountered in training}: target geometries other than the cube
(\cref{sec:results:gen:geometry}) and targets whose center of mass is displaced
from their geometric center (\cref{sec:results:gen:com}). The second is
\emph{extrapolation beyond the trained range} of the two quantities that govern
the difficulty of the task: the tumbling rate of the target
(\cref{sec:results:gen:tumble}) and its mass (\cref{sec:results:gen:mass}).
 
Each scenario is evaluated over $1000$ independent trials under the protocol of
\cref{sec:results:protocol}, with the same trained policy and the same success
criterion. In every scenario, exactly one factor is altered and the remainder are
held at their nominal distributions, so that any degradation is attributable to
the factor under test. \Cref{tab:generalization} reports, for every scenario,
both the outcome against the nominal reference and the corresponding constraint
satisfaction. Neither the alternative geometries nor the center-of-mass offsets
were sampled at any point during training, and the policy is evaluated on them
without retraining or fine-tuning.

\begin{table*}[t]
\centering
\caption{\textbf{Generalization and robustness}. Left: success rate, residual target velocities, and settling time, the latter three as means over successful trials. Right: constraint satisfaction, as the percentage of control steps at which each limit is exceeded; symbols are defined in \cref{tab:constraints}. The geometry and inertia scenarios are zero-shot; tumbling rates are per-axis bounds.}
\label{tab:generalization}
\small
\setlength{\tabcolsep}{3.5pt}
\begin{tabular}{@{}l cccc @{\quad} cccccc @{\quad} ccc @{\quad} cc@{}}
\toprule
& \multicolumn{4}{c}{\emph{Performance}} & \multicolumn{6}{c}{\emph{Mission constraints}} & \multicolumn{3}{c}{\emph{Actuation}} & \multicolumn{2}{c}{\emph{Safety}} \\
\cmidrule(lr){2-5} \cmidrule(lr){6-11} \cmidrule(lr){12-14} \cmidrule(lr){15-16}
Scenario & Succ. & $\lVert\bm{\omega}_t\rVert_f$ & $\lVert\bm{v}_t\rVert_f$ & $t_\mathrm{stab}$
& $c_{v_t}$ & $c_{\omega_t}$ & $c_{v_b}$ & $c_{\omega_b}$ & $c_{v_b}^{\,c}$ & $c_{\omega_b}^{\,c}$
& $c_{f}$ & $c_{\tau}$ & $c_{q}$ & $c_{F}$ & $c_{\mathrm{col}}$ \\
& (\%) & (rad/s) & (m/s) & (s) & \multicolumn{11}{c}{(\% of steps)} \\
\midrule
Nominal & 99.5 & 0.0036 & 0.0020 & 9.7 & 2.25 & 3.55 & 0.00 & 0.01 & 1.43 & 2.99 & 0.12 & 0.06 & 0.02 & 0.27 & \textbf{0.00} \\
\midrule
\multicolumn{16}{@{}l}{\emph{Unknown geometry} (zero-shot)}\\
\quad Cylinder & 99.2 & 0.0036 & 0.0021 & 6.5 & 2.24 & 3.83 & 0.13 & 0.11 & 3.00 & 4.51 & 0.19 & 0.06 & 0.03 & 0.22 & \textbf{0.00} \\
\quad Cuboid   & 99.7 & 0.0039 & 0.0021 & 7.7 & 2.22 & 4.08 & 0.04 & 0.04 & 2.74 & 5.03 & 0.22 & 0.08 & 0.03 & 0.25 & \textbf{0.00} \\
\midrule
\multicolumn{16}{@{}l}{\emph{Inertial asymmetry} (zero-shot)}\\
\quad CoM offset & 99.2 & 0.0039 & 0.0020 & 9.2 & 2.45 & 4.03 & 0.00 & 0.00 & 2.18 & 3.80 & 0.14 & 0.07 & 0.03 & 0.24 & \textbf{0.00} \\
\midrule
\multicolumn{16}{@{}l}{\emph{Tumbling rate} (per axis)}\\
\quad $0.35$~rad/s & 99.0 & 0.0048 & 0.0023 & 9.3 & 2.91 & 5.86 & 0.05 & 0.06 & 6.48 & 7.68 & 0.26 & 0.14 & 0.03 & 0.32 & \textbf{0.00} \\
\quad $0.45$~rad/s & 95.3 & 0.0065 & 0.0026 & 11.5 & 5.08 & 13.00 & 0.39 & 0.50 & 13.10 & 18.71 & 0.87 & 0.43 & 0.03 & 0.36 & \textbf{0.00} \\
\quad $0.55$~rad/s & 88.7 & 0.0083 & 0.0032 & 13.0 & 8.85 & 20.50 & 1.46 & 2.31 & 21.57 & 28.49 & 2.04 & 1.02 & 0.09 & 0.39 & \textbf{0.00} \\
\quad $0.65$~rad/s & 79.8 & 0.0093 & 0.0036 & 14.4 & 13.20 & 26.79 & 3.12 & 5.14 & 28.72 & 37.65 & 3.36 & 1.72 & 0.20 & 0.45 & \textbf{0.00} \\
\midrule
\multicolumn{16}{@{}l}{\emph{Target mass}}\\
\quad $60$--$75$~kg   & 99.6 & 0.0036 & 0.0021 & 6.0 & 1.72 & 4.13 & 0.02 & 0.02 & 3.37 & 4.67 & 0.16 & 0.06 & 0.04 & 0.28 & \textbf{0.00} \\
\quad $75$--$90$~kg   & 99.5 & 0.0037 & 0.0021 & 5.9 & 1.57 & 4.46 & 0.04 & 0.05 & 3.84 & 5.07 & 0.21 & 0.08 & 0.05 & 0.31 & \textbf{0.00} \\
\quad $90$--$120$~kg  & 99.8 & 0.0039 & 0.0021 & 5.9 & 1.39 & 5.00 & 0.07 & 0.06 & 4.64 & 6.10 & 0.30 & 0.11 & 0.06 & 0.34 & \textbf{0.00} \\
\quad $120$--$180$~kg & 99.6 & 0.0042 & 0.0022 & 4.7 & 1.15 & 5.51 & 0.18 & 0.14 & 6.77 & 7.73 & 0.52 & 0.17 & 0.09 & 0.35 & \textbf{0.00} \\
\quad $180$--$300$~kg & 99.2 & 0.0048 & 0.0023 & 4.0 & 0.97 & 7.49 & 0.26 & 0.33 & 9.55 & 11.48 & 0.76 & 0.33 & 0.13 & 0.39 & \textbf{0.00} \\
\bottomrule
\end{tabular}
\end{table*}
\subsubsection{Unknown Geometry}
\label{sec:results:gen:geometry}
 
\begin{figure}[t]\centering
  \begin{subfigure}{0.48\linewidth}
    \includegraphics[width=\linewidth]{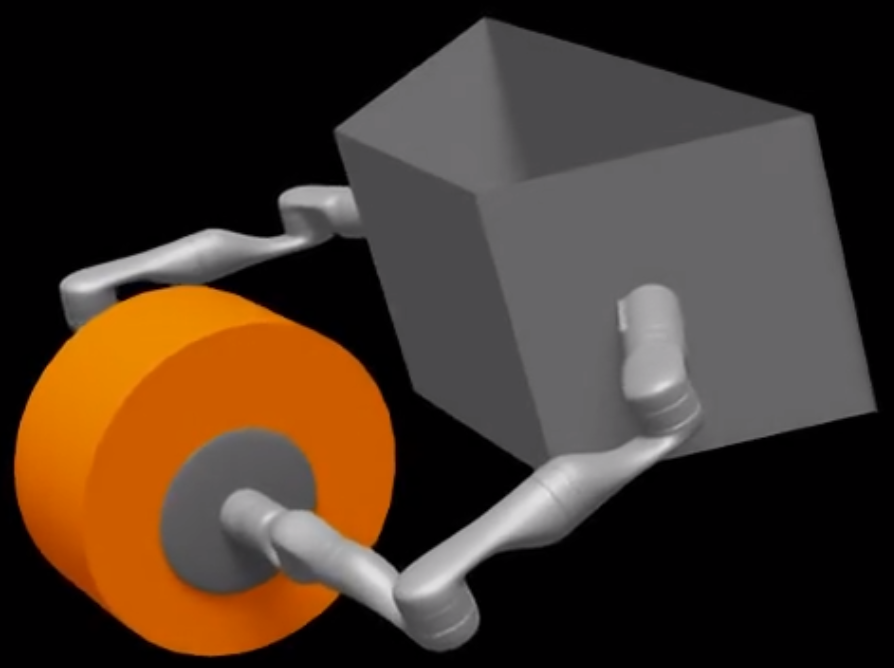}
    \caption{}
    \label{fig:gen_cylinder}
  \end{subfigure}
  \begin{subfigure}{0.48\linewidth}
    \includegraphics[width=\linewidth]{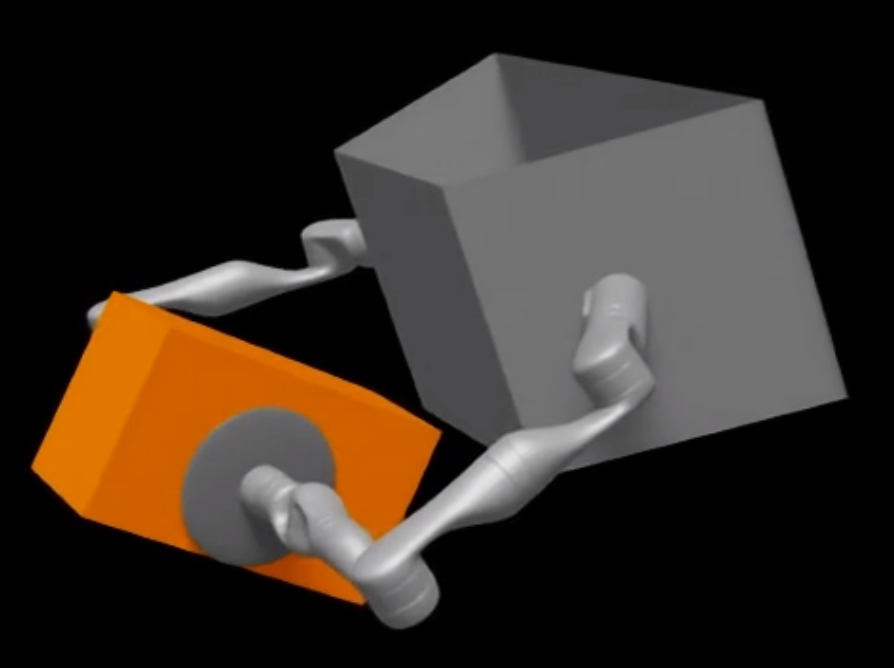}
    \caption{}
    \label{fig:gen_cuboid}
  \end{subfigure}
  \caption{\textbf{Zero-shot capture of unseen geometries}. The policy is unchanged from the cube-trained one.}
  \label{fig:gen_geometry}
\end{figure}

The capture strategy of \cref{sec:method} is fixture-free: the end effectors establish a friction grasp by pressing on opposing faces. Whether that competence depends on the cube's flat, equal faces is tested by evaluating the trained policy on two geometries never sampled during training: a cylinder
of diameter $0.5~\mathrm{m}$ and height $0.25~\mathrm{m}$ (\cref{fig:gen_cylinder}), and a rectangular
cuboid of $0.3\times0.25\times0.5~\mathrm{m}$ (\cref{fig:gen_cuboid}), both carrying roughly $2.2$ times
the mean moment of inertia of the trained cube at equal mass. The policy network
is unchanged; only the width parameter that places the end-effector command
targets in \cref{sec:method:command} is set to the known extent of the new body,
so the transfer is zero-shot at the level of the policy. 

The cylinder succeeds in $99.2\%$ of trials and the cuboid in $99.7\%$, against $99.5\%$ on the cube. More telling is the phase breakdown: on both geometries the servicer completes the approach and establishes a bilateral friction grasp in \emph{every} trial, seating the end effectors on a continuously curved surface, and on faces of an extent and separation never seen in training, as reliably as on the cubic faces. Safety is likewise unaffected, with no collision in any trial and the contact-force constraint violated marginally less often than nominally.
 
One difference does appear. Over both geometries the contact-gated
base-velocity constraints are violated roughly twice as often as nominally,
consistent with a smaller and less predictable contact patch transmitting the
reaction of the detumbling couple less cleanly than a flat face of known extent.
The single loss of grasp observed across these two scenarios occurs on the curved geometry, where that patch is smallest.

\subsubsection{Unmodeled Inertial Asymmetry}
\label{sec:results:gen:com}
 
Assumption~\ref{as:inertia} states that the inertial parameters of the target are
unknown, and the method of \cref{sec:method} performs no identification of them.
Randomizing the target mass tests only part of that assumption: it scales the
inertia tensor without altering its structure, and leaves the center of mass at
the geometric center. The more demanding case is a displaced center of mass,
since the servicer observes only the pose of the body, so a wrench applied on the
geometric faces exerts an unpredictable torque about the true center of mass,
coupling the target's translational and rotational responses in a way they are
not for a symmetric body. This is evaluated by displacing the center of mass by
an offset drawn uniformly from $\pm0.05~\mathrm{m}$ on each axis, giving
displacements of up to $0.087~\mathrm{m}$: $69\%$ of the cube's half-side and
$72\%$ of the pressing moment arm.
 
The effect on the outcome is small. The policy succeeds in $99.2\%$ of trials
against $99.5\%$ nominally, and completes the approach and establishes a
bilateral grasp in every trial. What is most informative is how little the
remaining statistics move: the residual target and base velocities are
indistinguishable from the nominal values at every reported digit, the contact
forces are unchanged, and the contact-gated base-velocity constraints rise only
moderately. No new failure mode appears, and none involves a loss of contact or a
safety violation.
 
These results support the claim that the policy accommodates unknown inertial
parameters by feedback rather than by identification. It regulates contact
forces and target velocity closely enough that a substantial and unobservable
inertial asymmetry is absorbed as a disturbance, without any estimate of where
the center of mass lies. This distinguishes the approach from
model-based capture strategies, which require the target's inertial parameters
before a stabilizing controller can be synthesized.

\subsubsection{Extrapolation in Tumbling Rate}
\label{sec:results:gen:tumble}
 
During training each component of $\bm{\omega}_t$ is drawn uniformly from
$\pm0.25~\mathrm{rad}\,\mathrm{s}^{-1}$, so the \emph{magnitude} of the initial
tumbling rate has mean $0.24~\mathrm{rad}\,\mathrm{s}^{-1}$ and cannot exceed
$0.25\sqrt{3} \approx 0.43~\mathrm{rad}\,\mathrm{s}^{-1}$. Four further batches
raise the per-component bound to $0.35$, $0.45$, $0.55$ and
$0.65~\mathrm{rad}\,\mathrm{s}^{-1}$, all else nominal. Because each batch
samples a broad range of magnitudes rather than a single value, the results are
reported as a function of the rate actually realized in each trial: all batches
are pooled and binned by $\|\bm{\omega}_t(0)\|$, which spans $0.15$ to
$0.90~\mathrm{rad}\,\mathrm{s}^{-1}$ over the four batches.
 
\begin{figure}[t]\centering
  \includegraphics[width=0.8\linewidth]{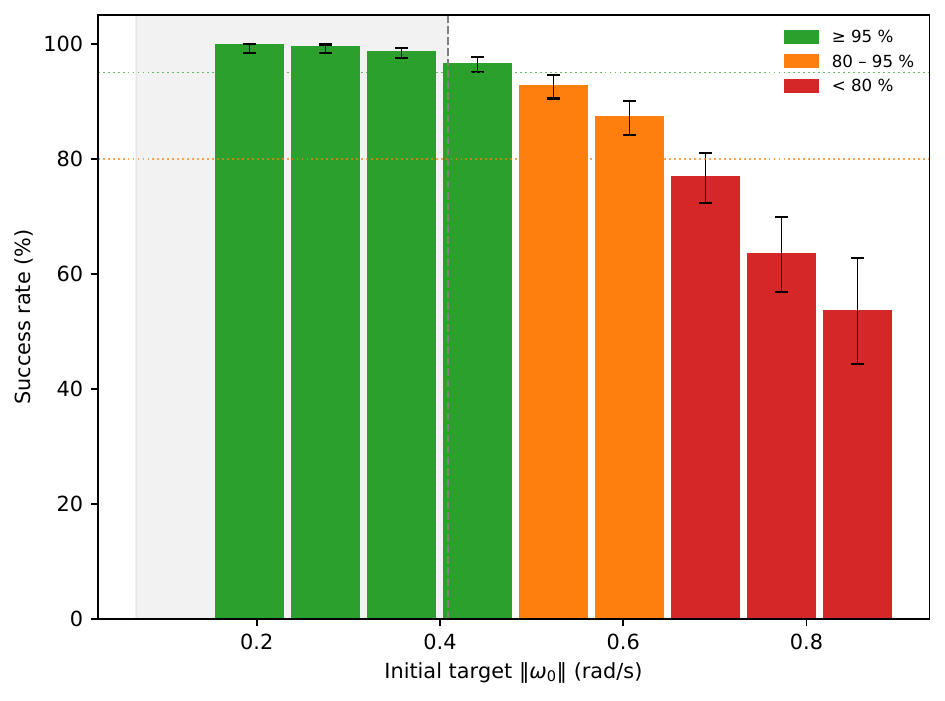}
  \caption{\textbf{Success rate versus the initial tumbling magnitude}. Error bars are $95\%$ Wilson intervals; the shaded region marks
  magnitudes attainable during training.}
  \label{fig:gen_tumble}
\end{figure}

\Cref{fig:gen_tumble} shows the result. Within the trained range the policy is
unaffected, at $100.0\%$, $99.6\%$ and $98.7\%$ in the three bins below
$0.43~\mathrm{rad}\,\mathrm{s}^{-1}$. Beyond it, performance declines smoothly
and monotonically, with no discontinuity at the boundary of the training
distribution: success remains above $95\%$ out to
$0.48~\mathrm{rad}\,\mathrm{s}^{-1}$, twice the mean rate seen in training; above
$87\%$ out to $0.65~\mathrm{rad}\,\mathrm{s}^{-1}$, $1.5$ times the fastest rate
attainable in training; and above one half out to
$0.90~\mathrm{rad}\,\mathrm{s}^{-1}$, more than twice that ceiling. The policy
does not fail abruptly when the tumbling rate leaves the distribution it was
trained on; it degrades in proportion to how far outside that distribution the
target lies. Safety is maintained throughout: across all four elevated-rate
batches the base-collision constraint is never violated, and the end-effector contact-force constraint is exceeded in at most $0.45\%$ of steps. What degrades outside the training distribution is the ability to complete the mission, not the ability to remain safe while attempting it.
 
The composition of the failures is more informative than the rate itself, because
it changes character as the tumbling rate rises. In the nominal case, and in the
slowest of the four batches, \emph{every} failure is of the residual-precision
type analyzed in \cref{sec:results:failure}: the grasp is established and
retained, and only the final angular velocity misses the tolerance. As the rate
increases, failures of the grasp itself appear and come to dominate, rising from
$0\%$ of failures nominally to $38.6\%$ in the fastest batch, while the rate at
which a bilateral grasp is established at all falls from $100\%$ to $97.8\%$.
This identifies what limits the method at high tumbling rates, and it is not the
removal of angular momentum. The contact faces translate tangentially at a speed
proportional to $\|\bm{\omega}_t\|$, so the manipulators must track a
faster-moving surface in order to seat and to maintain the grasp. The steady-state normal force falls from $11.0 \pm 0.3~\mathrm{N}$ to
$9.9 \pm 2.8~\mathrm{N}$ over the same interval, the increase in variance
indicating that contact is intermittently lost and re-established rather than
held steadily.

\subsubsection{Extrapolation in Target Mass}
\label{sec:results:gen:mass}
 
Target mass is randomized during training over $15$--$45~\mathrm{kg}$. Five
further batches raise it to $60$--$75$, $75$--$90$, $90$--$120$, $120$--$180$ and
$180$--$300~\mathrm{kg}$; the heaviest exceeds the
$150~\mathrm{kg}$ mass of the servicer itself and represents a tenfold increase
over the nominal target. Since the tumbling-rate distribution is unchanged, the
angular momentum to be removed grows in proportion.
 
The success rate does not degrade, remaining between $99.2\%$ and $99.8\%$ across
the five batches against $99.5\%$ nominally, including where the target outweighs
the servicer by up to a factor of two. Detumbling is evidently limited by the
precision to which the residual must be nulled rather than by the authority
available to remove angular momentum: settling is in fact reached \emph{sooner}
as the target grows heavier, the mean time falling from $9.7~\mathrm{s}$
nominally to $4.0~\mathrm{s}$ in the heaviest batch, a larger inertia being less
responsive to the small contact disturbances that delay satisfaction of the dwell
criterion in the light-target case.
 
What does degrade with mass is the quiescence of the servicer base. The
contact-gated base-velocity constraints rise from $1.4\%$ and $3.0\%$ of steps
nominally to $9.6\%$ and $11.5\%$ in the heaviest batch, and the $\Delta v$
expended rises from $0.045$ to $0.085~\mathrm{m}\,\mathrm{s}^{-1}$. The servicer
must reject the reaction of the detumbling couple using a propulsion authority
that is fixed, while the momentum it reacts against grows by an order of
magnitude; below the mass of the servicer that reaction is absorbed comfortably,
and above it the target is the heavier body and the base is displaced more in the
exchange. This is a limit imposed by the propulsion authority of the platform
rather than a failure of the policy. Read against \cref{sec:results:gen:tumble}, these results identify which property of the target governs the difficulty of fixture-free capture: a tenfold increase in the momentum to be removed leaves the success rate unchanged and the settling time shorter, whereas a factor of $2.6$ increase in the speed at which the contact surface travels costs twenty points of success. The binding constraint is the bandwidth with which the manipulators can track and hold a moving contact patch, not the inertia of the body they are holding.


\subsection{Ablations}
\label{sec:results:ablations}

The design of \cref{sec:method} places the proximity gate at three loci of the
CMDP. This subsection varies each and evaluates the resulting policies under the
protocol of \cref{sec:results:protocol}. Removing the reward gating replaces the
complementary weighting of \cref{sec:method:reward} with a plain sum of the same
terms; removing the action gating deletes $g_p$ from \eqref{eq:action_gate}, so
that joint increments are applied in full at any separation. The constraint
decomposition of \cref{sec:method:cat_gating} is replaced not by an
unconstrained formulation but by the natural single-constraint alternative, in
which one constraint per axis carries a limit interpolated between the loose and
tight values according to the contact signal, $\bar{v} =
v^{\mathrm{loose}}(1-\mathbb{1}_{c}) + v^{\mathrm{tight}}\mathbb{1}_{c}$, a
variant that is phase-aware in every respect except one, maintaining a single
normalization scale where the decomposition maintains two. All configurations
were trained with an identical budget and hyperparameters and evaluated over
$1000$ nominal trials (\cref{tab:ablations}).

The two gating loci prove partially redundant but jointly necessary. With the decomposition retained, removing either alone costs little: $97.3\%$
without action gating and $92.1\%$ without reward gating, against $99.5\%$ for
the full design. Removing both reduces success to $54.6\%$. Either locus
therefore suffices to induce the phase separation the task depends on. They are
not equivalent, however: every configuration lacking reward gating expends an
order of magnitude more propulsive effort ($0.71$ to $0.73$ against
$0.045~\mathrm{m}\,\mathrm{s}^{-1}$) and regulates the press markedly less well
($8.7$ to $8.8~\mathrm{N}$ steady contact force against $11.0~\mathrm{N}$). The
reward gate is what produces an efficient, well regulated capture; the action
gate provides, in addition, a guarantee that holds irrespective of what the
policy has learned. Their combined absence is also the only condition under which
the grasp itself becomes unreliable: among the configurations retaining the
decomposition, contact is lost before the end of the episode in $10.8\%$ of
trials when both gates are removed, and in none whenever either is present.

The constraint decomposition proves the least substitutable of the three.
Replacing it with the interpolated single-limit constraint reduces success from
$99.5\%$ to $23.8\%$, the lowest of any configuration tested, and costs
performance at every level of gating: with neither gate present, success falls
from $54.6\%$ to $36.9\%$. The interpolated variant fails neither at the
approach, which it completes in every trial, nor at the grasp, which it
establishes in every trial and retains in all but $3.5\%$; it fails at the
residual, leaving the base at $0.0137~\mathrm{m}\,\mathrm{s}^{-1}$ and
$0.0149~\mathrm{rad}\,\mathrm{s}^{-1}$ against $0.0028$ and $0.0040$ for the full
design. This is the dilution predicted in \cref{sec:method:cat_gating}: tightening a limit during contact is not sufficient
by itself, because a single constraint carries a single normalization scale set
by the approach transient. The limit must be carried by a separate constraint, so
that it is normalized against the violations it is intended to suppress.
\begin{table}[t]
\centering
\caption{\textbf{Ablations}. R and A denote the reward and action-space gating of
\cref{sec:method:reward,sec:method:action}; C denotes the constraint formulation
of \cref{sec:method:cat_gating}, either the decomposed loose and tight pair
(\emph{dec.}) or the single constraint whose limit is interpolated between the
same two values according to the contact signal (\emph{int.}). Grasp loss is the
percentage of trials in which contact is established and subsequently lost.
Residuals are means over successful trials.}
\label{tab:ablations}
\setlength{\tabcolsep}{3pt}
\begin{tabular}{cccccccc}
\toprule
R & A & C & Success & Grasp loss & $\lVert\bm{\omega}_t\rVert_f$ & $\lVert\bm{\omega}_b\rVert_f$ & $\Delta v$ \\
 & & & (\%) & (\%) & (rad/s) & (rad/s) & (m/s) \\
\midrule
\checkmark & \checkmark & dec. & \textbf{99.5} & 0.0 & 0.0036 & 0.0040 & 0.045 \\
\midrule
\checkmark & --         & dec. & 97.3 & 0.0  & 0.0087 & 0.0058 & 0.189 \\
--         & \checkmark & dec. & 92.1 & 0.0  & 0.0154 & 0.0115 & 0.727 \\
--         & --         & dec. & 54.6 & 10.8 & 0.0433 & 0.0296 & 0.707 \\
\midrule
\checkmark & \checkmark & int. & 23.8 & 3.5  & 0.0311 & 0.0149 & 0.254 \\
--         & --         & int. & 36.9 & 37.3 & 0.0963 & 0.0371 & 0.733 \\
\bottomrule
\end{tabular}
\end{table}


\subsection{Comparison with Model-Based Methods}
\label{sec:results:comparison}

\begin{table*}[t]\centering
\caption{Comparison with representative model-based and hybrid capture/detumbling
methods, along the assumptions each requires. Reported outcomes are quoted under
each method's own conditions and are not comparable across rows.}
\label{tab:baselines}
\small
\setlength{\tabcolsep}{4pt}
\begin{tabular}{@{}p{2.6cm}cccccp{2.9cm}@{}}
\toprule
Method & \makecell{Target\\state} & \makecell{Inertial\\params.} & \makecell{Phase\\structure} & \makecell{Constraints\\in loop} & \makecell{Fixture-\\free} & Reported outcome \\
\midrule
Friction detumbling \cite{chen2020friction}
  & full & required & switched & no & yes
  & $20^\circ$/s halted in $30$~s, base $<0.15^\circ$ \\
Adaptive detumbling \cite{zhou2023adaptive}
  & full & online ID & post-capture & no & yes
  & inertia uncertainty rejected \\
Robust SMC \cite{wang2023robust}
  & full & bounded & switched & no & no
  & chatter-free second-order SMC \\
Time-optimal \cite{li2024timeoptimal}
  & full & bounded & plan+track & bounds & no
  & detumble and pose in one action \\
Optimal concurrent \cite{xi2024adaptive}
  & full & online ID & plan+track & \makecell{actuator\\+motion} & yes
  & finite-time, low chatter \\
Position/force \cite{wang2024positionforce}
  & partial & none & \makecell{switched\\SMC+RL} & joints & yes
  & PPO force loop, unknown target \\
\textbf{Ours}
  & \textbf{noisy rel.} & \textbf{none} & \textbf{single policy}
  & \textbf{all, cont.} & \textbf{yes}
  & \textbf{$99.5\%$/$1000$, end-to-end} \\
\bottomrule
\end{tabular}
\end{table*}

This subsection places the preceding results in the context of the model-based
and hybrid methods that represent the current state of the art in dual-arm
capture and detumbling. A direct numerical comparison is not meaningful: none of
these methods shares the servicer model, target set, success criterion, or
simulator used here, and none has released an implementation, so any reproduction
would compare against a reconstruction rather than the method itself. The
comparison is therefore drawn along the axis a shared benchmark would obscure:
the assumptions each method requires in order to operate
(\cref{tab:baselines}).

Three such assumptions recur in the literature and are absent from the present
method. The first is knowledge of the target's inertial parameters.
Friction-based detumbling regulates the contact force against a target whose
inertia is taken as known \cite{chen2020friction}; where it is not, it must be
identified online or compensated adaptively \cite{zhou2023adaptive,
xi2024adaptive}, and in every such case the estimate converges only after a
dedicated excitation transient, during which the stabilizing controller cannot
yet act on a correct model. The present policy performs no identification:
\cref{sec:results:gen:com} shows it absorbs a center-of-mass displacement of up
to $72\%$ of the pressing moment arm, entirely unmodeled, with no measurable
change in residual precision.

The second is a hand-designed decomposition of the mission into phase-specific
controllers joined by switching logic. Robust and time-optimal formulations plan
or track a trajectory for a single phase and rely on separate laws for the others
\cite{wang2023robust, li2024timeoptimal}, and even the hybrid learning method
closest to this work composes a sliding-mode motion controller with a separately
trained force loop \cite{wang2024positionforce}. The switching instant coincides
with the contact discontinuity, which \cref{sec:intro} identifies as the point of
greatest fragility; \cref{sec:results:emergent} shows the present policy crossing
that boundary without a controller handoff and without an impulsive disturbance
to the base, because the phase structure is internal to a single continuous
policy rather than imposed by a switch.

The third is the treatment of safety limits. Where actuator or target-motion
bounds are considered at all, they enter as constraints on a planned trajectory
\cite{li2024timeoptimal, xi2024adaptive} rather than as limits enforced on the
closed-loop controller at every step and across every phase.
\Cref{sec:results:constraints} shows the present policy holding contact-force,
actuator, base-reaction, and collision limits continuously through the entire
mission, and \cref{sec:results:ablations} shows that the constraint decomposition
is what secures this.

What the comparison establishes is therefore not a performance ranking, but that
the methods in the upper rows of \cref{tab:baselines} each require target-model knowledge, an identification phase, or a designed
phase switch, and most require all three. The present method requires none, while enforcing its safety limits in the closed loop rather than in a plan.

\section{Conclusions}
\label{sec:conclusions}

In this paper, we show that a single reinforcement-learning policy can execute the complete
active-debris-removal mission end-to-end, with no grappling fixture, no
identification of the target's inertial parameters, and no hand-designed switch
between phases. Observing only noisy relative state and its own
proprioception, the policy drives a dual-arm free-flying servicer through the
entire sequence, succeeding in $99.5\%$ of $1000$ randomized trials while holding
its contact-force, actuator, base-reaction, and collision constraints throughout.

The enabling idea is a single proximity gate applied at three loci of a CMDP. At
the reward, the gate and its complement form a partition of unity interpolating
between the opposed objectives of closing and capture, so that the phase
structure is induced by the task rather than imposed by a supervisor. At the action space, the same gate suppresses manipulator authority at range.
This forecloses the grab-and-pull exploit available to a free-flying servicer
whose target is the lighter body, and it does so as a hard guarantee rather than
a learned preference. At the constraint set, we identified and resolved a normalization
failure of CaT in multi-phase tasks: a constraint spanning phases that load it at different magnitudes has its normalization captured by the most violated phase, and is rendered inert where
precision matters most. Decomposing it into an always-active loose term and a contact-gated tight term restores its authority. This failure mode and its remedy apply to any multi-phase CRL problem that aims to use CaT.

The evaluation supports these claims on several independent axes. The policy transfers zero-shot to target geometries and centre-of-mass offsets never seen in training. It extrapolates across a tenfold range of target mass, up to a target twice the mass of the servicer, with no loss of success and a shorter settling time. When the tumbling rate is pushed beyond twice the fastest
rate seen in training, it degrades gracefully rather than abruptly. Read together, these
studies localize the binding difficulty of fixture-free capture: it is set by the
bandwidth with which the manipulators can track and hold a moving contact patch,
not by the inertia of the body they hold. Ablations confirm that each gating locus contributes and that the constraint decomposition is necessary: its natural single-constraint alternative fails as
the normalization analysis predicts.

Several limitations bound these conclusions. The results are obtained in
simulation; a hardware demonstration is the necessary next step, and the
sim-to-real gap for the contact dynamics on which the method depends is not
characterized here. The residual failures, though benign, are not fully explained, and finer
residual control, whether through a third contact point or momentum-aware
placement of the two, is left to future work. The target is
treated as a single rigid body, so flexible appendages, propellant slosh, and
multi-body debris are beyond the present scope. Finally, the method enforces its
constraints with high but not certified reliability: violations are driven to a
small fraction of control steps and the hard collision constraint was never
violated across the trials reported, but no worst-case guarantee is provided, and
coupling learned capture with formal safety certification is an important
direction.

ADR demands phase structure, safety enforcement, and robustness to unknown
targets. These results indicate that a single continuously gated policy, trained
under constraints, can acquire all three at once, rather than assembling them
from separately designed and separately switched components. The gating and
constraint-decomposition mechanisms apply to multi-phase, contact-rich,
constrained control problems beyond the debris-removal setting studied here.



\bibliographystyle{IEEEtran}
\bibliography{bibliography}

\end{document}